\RequirePackage{fix-cm}
\documentclass[twocolumn]{svjour3arxiv}          
\smartqed  
\usepackage{graphicx}
 \usepackage{mathptmx}      
\usepackage{amsmath}
\usepackage{amsfonts}
\usepackage{bm}
\usepackage{color}
\usepackage{subfig}
\usepackage{natbib}
\usepackage[labelfont=bf]{caption}
\DeclareMathOperator{\erf}{erf}

\DeclareMathOperator{\rank}{rank}
\DeclareMathOperator{\diag}{diag}
\DeclareMathOperator{\vecrow}{Vec}
\DeclareMathOperator{\vecaff}{Vec_6}

\DeclareMathOperator*{\argmin}{argmin}
\DeclareMathOperator*{\argmax}{argmax}

\journalname{}
\begin{document}

\title{Active Annotation of Informative Overlapping Frames in Video Mosaicking Applications}
\titlerunning{Active Annotation of Informative Overlapping Frames in Video Mosaicking Applications}

\author{Lo\"ic Peter \and Marcel Tella-Amo \and Dzhoshkun Ismail Shakir \and Jan Deprest \and \\  S\'ebastien Ourselin  \and Juan Eugenio Iglesias \and Tom Vercauteren}


\institute{Lo\"ic Peter, Marcel Tella-Amo \at Wellcome / EPSRC Centre for Interventional and Surgical Sciences, University College London, United Kingdom
\and
Lo\"ic Peter, Juan Eugenio Iglesias \at Centre for Medical Image Computing, University College London, United Kingdom
\and
Juan Eugenio Iglesias \at Martinos Center for Biomedical Imaging, Massachusetts General Hospital and Harvard Medical School
\and
Juan Eugenio Iglesias \at Computer Science and Artificial Intelligence Laboratory (CSAIL), Massachusetts Institute of Technology
\and
Jan Deprest, Tom Vercauteren \at KU Leuven, Belgium
\and
Dzhoshkun Ismail Shakir, Jan Deprest, S\'ebastien Ourselin, Tom Vercauteren \at King's College London, London, United Kingdom
}

\date{}

\maketitle

\begin{abstract}

Video mosaicking requires the registration of overlapping frames located at distant timepoints in the sequence to ensure global consistency of the reconstructed scene. However, fully automated registration of such long-range pairs is (\textit{i})~challenging when the registration of images itself is difficult; and (\textit{ii})~computationally expensive for long sequences due to the large number of candidate pairs for registration. In this paper, we introduce an efficient framework for the active annotation of long-range pairwise correspondences in a sequence. Our framework suggests pairs of images that are sought to be informative to an oracle agent (e.g., a human user, or a reliable matching algorithm) who provides visual correspondences on each suggested pair. Informative pairs are retrieved according to an iterative strategy based on a principled annotation reward coupled with two complementary and online adaptable models of frame overlap. In addition to the efficient construction of a mosaic, our framework provides, as a by-product, ground truth landmark correspondences which can be used for evaluation or learning purposes. We evaluate our approach in both automated and interactive scenarios via experiments on synthetic sequences, on a publicly available dataset for aerial imaging and on a clinical dataset for placenta mosaicking during fetal surgery.



\keywords{Image Mosaicking \and Active Learning \and Bundle Adjustment \and Human-Computer Interaction \and Fetoscopy}
\end{abstract}

\section{Introduction}
\label{sec:introduction}

The purpose of video mosaicking is to reconstruct a scene based on a video sequence in which each frame only provides an incomplete local view of the scene of interest. A well-known example of mosaicking task is the panoramic image stitching found in standard digital cameras~\citep{Brown2007}. Mosaicking algorithms are particularly needed in a variety of biomedical applications such as retinal imaging~\citep{can2002pami,choe2005registration,prokopetc2016reducing,richa2014tmi}, fetoscopy~\citep{Reeff,tella2018probabilistic}, fibered endoscopy~\citep{Atasoy2008}, confocal endomicroscopy~\citep{loewke2011vivo,mahe2013viterbi,VERCAUTEREN2006media} and 3D ultrasound imaging~\citep{ni2008volumetric,wachinger2007three}. Applications of video mosaicking are also typically found in remote sensing applications, for example to reconstruct the seabed from underwater sequences~\citep{Gracias2004ifac,Elibol2014osa} or to obtain a high-resolution map of a city from a set of aerial images~\citep{Kekec2014ras,molina2014persistent,Xia2017pr}.

\begin{figure}[!t]
\centering
\subfloat[Frame \#1
]{\includegraphics[width=0.15\textwidth]{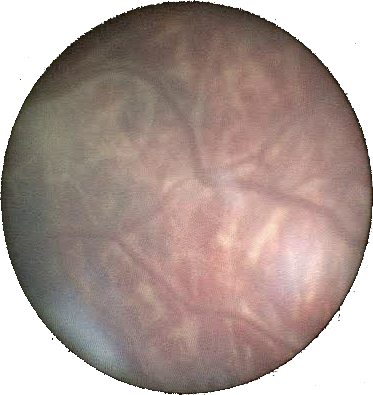}}\hfill
\subfloat[Frame \#243
] {\includegraphics[width=0.15\textwidth]{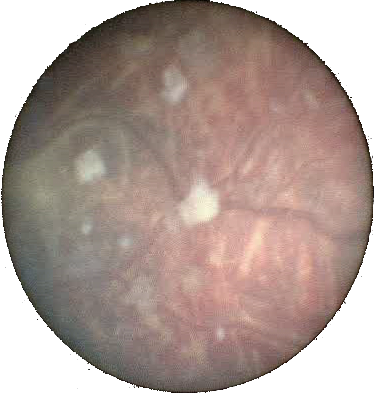}}\hfill
\subfloat[Frame \#459
]{\includegraphics[width=0.15\textwidth]{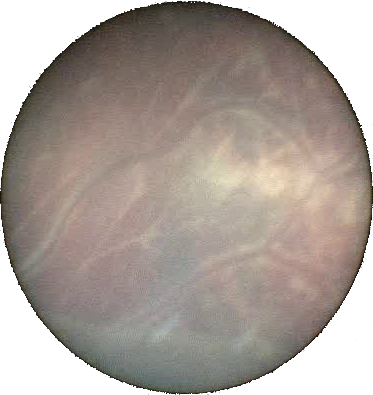}}
\caption{\textbf{Challenging long-range overlapping frames.} These three frames represent the same location at different timepoints of a video sequence of $600$ frames acquired during in vivo fetoscopy. The differences in terms of visual conditions (artifacts or perspective changes) make a reliable automated registration of these different views challenging.}
\label{fig:example_low_range}
\end{figure}

Mosaicking approaches typically build on the pairwise registration of pairs of frames of the sequence which spatially overlap (such as frames acquired at consecutive timepoints). A consistent mosaic is then created as to achieve a compromise between these multiple pairwise registrations of overlapping frames, e.g. via bundle adjustment~\citep{szeliski2007image,triggs2000ba}. To ensure the global consistency of the reconstructed panorama and avoid the accumulation of registration errors, it is crucial to also register frames that are at distant timepoints in the sequence and yet spatially overlap, i.e. corresponding to cases where the camera trajectory revisits a part of the scene. Unfortunately, the feasibility to do so critically depends on (\textit{i}) the identification of long-range overlapping frames of the sequence, a task called \textit{topology inference} in the literature~\citep{Sawhney1998eccv}, and (\textit{ii}) the \textit{accurate and reliable registration} of these long-range pairs. Satisfying these two objectives in a tractable and consistent manner is particularly difficult in cases of long sequences and in challenging visual conditions in which highly distinct keypoints cannot be found and tracked, such as with low-resolution biomedical data (Fig.~\ref{fig:example_low_range}).

In addition, the necessity of inferring the topology of the sequence accurately complicates the creation of a gold standard for a given sequence.
In the case of pairwise image registration, algorithms can be evaluated by measuring the error made by the predicted transformation on a set of visual correspondences that were manually identified within the two registered images~\citep{murphy2011semi}. Adopting such a landmark-based strategy in a mosaicking context would require a much higher annotation cost, as one would ideally need to annotate each pair of overlapping images in the sequence to quantitatively estimate the quality of a reconstructed mosaic. To achieve a lower annotation cost, a reasonable compromise could be to annotate landmarks of the scene that are observed at various timepoints of the sequence and that are as representative as possible of its topology, given a fixed annotation budget. However, deciding on the most relevant subset of image pairs to annotate is not only subjective when performed manually, but also highly tedious and challenging for long sequences with complex trajectories. Indeed, finding annotable pairs requires to identify and remember revisited locations in the sequence,  a task which is far from trivial in visually challenging environments. For these reasons, mosaicking algorithms are notoriously difficult to evaluate quantitatively~\citep{Atasoy2008,prokopetc2016reducing}.
Apart from synthetic and phantom experiments for which a gold standard is more easily available, mosaicking algorithms in ``real-world" scenarios are often only evaluated by qualitative visual assesssment of the rendered mosaic~\citep{Atasoy2008,mahe2015motion}, or by using as a surrogate gold standard the result of a more computationally intensive but more reliable algorithm such as bundle adjustment~\citep{Kekec2014ras}. The feasibility of the second option is, however, naturally limited in challenging cases where out-of-the-box bundle adjustment is not sufficiently accurate. This situation is, again, frequent when the pairwise registration of long-range frames and the estimated topology of the sequence are not fully reliable.

In this paper, we introduce an active annotation framework enabling the identification and registration of long-range overlapping pairs in a sequence. Starting from a chain of registrations between consecutive frames of the sequence, our approach iteratively retrieves pairs of frames that are sought to be informative on which an agent, acting as an oracle, provides annotations. The annotations provided by the agent consist of a binary label indicating the presence or absence of overlap between the two suggested frames and, in the former case, of a set of pointwise correspondences relating these two frames. A fundamental assumption of our scenario is that the agent gives reliable but costly annotations. As a result, it is crucial to minimise the number of queries made to the agent. To do so, we introduce a principled retrieval strategy where each suggested pair is chosen to satisfy a compromise between: (\textit{i}) its informativeness, i.e. its non-redundancy with the previously annotated pairs; and (\textit{ii}) the probability that the suggested pair is indeed annotable, i.e. that the two frames spatially overlap.

Two types of annotating agent can be considered. First, the agent can be a human user who, depending on the scenario, either directly annotates the retrieved pair in an interactive fashion, or visually controls the output of a registration algorithm and corrects it if needed. In this case, the associated human effort naturally imposes the relevance of only making sparse, well-chosen queries to the user. Alternatively, the agent can also be an automated, reliable yet possibly computationally intensive registration algorithm. To consider such an agent, the registration algorithm must not only be accurate, but also able to identify its failure modes and its inability to register the two frames (for example, if they do not overlap). In this case, our framework seeks to achieve a well-known objective in the literature~\citep{Elibol2013ras,Sawhney1998eccv,tella2019pruning,Xia2017pr}, namely the minimisation of the amount of matching attempts, and thus the computational complexity of the reconstruction. The applicability of our framework is twofold: the provided annotations can not only be leveraged as reliable correspondences to build the mosaic, but also be stored as well-chosen informative ground truth landmarks on this sequence for further experimentation.

We evaluate our framework in both contexts: on a publicly available sequence for aerial imaging~\citep{Xia2017pr} for which an automated reliable pairwise registration algorithm is available, and on a medical sequence for the mosaicking of placenta in in vivo fetoscopy~\citep{Peter2018ijcars} for which human inspections are beneficial. In addition, we evaluate our framework on controlled synthetic examples to illustrate its main fundamental properties. We compare our approach with existing methods for topology inference and perform ablation studies to investigate its behaviour. Our experimental validation demonstrates the ability of our method to effectively reduce the number of queries, and thus the number of long-range registration, to achieve a given mosaic recostruction accuracy.

%

The outline of the paper is as follows. In Section~\ref{sec:image_mosaicking}, we formalise the problem of video mosaicking and highlight three frequent challenges arising when combining multiple pairwise registrations. In Section~\ref{sec:related_work}, we review the related work on mosaicking with a particular emphasis on these three challenges. We give an overview of our methodological contributions in Section~\ref{sec:contributions} before describing our methodology in details in Section~\ref{sec:methods}. Finally, after presenting in Section~\ref{sec:baselines} the baselines used in our experimental evaluation, we present our experiments and results in Section~\ref{sec:experiments}.

\section{Image Mosaicking from Multiple Pairwise Correspondences}
\label{sec:image_mosaicking}

Before presenting the related work, we pose the general problem of video mosaicking on which most of the prior art is based. We consider a sequence of $N$ images $(I_n)_{1 \leq n \leq N}$, where each $I_n$ offers an incomplete view of a scene to be reconstructed. Each image $I_n$ is defined over a domain $\Omega_n$ corresponding to the 2D camera space, which we assume for simplicity to be the same domain $\Omega$ for each image. A usual example of a domain $\Omega$ is a rectangle of fixed size determined by the resolution of the camera sensor. Mosaicking applications (in contrast to 3D reconstruction) are typically characterised by the imaging of a static planar or quasi-planar scene, where the imaged scene has relatively large dimensions in comparison to the field-of-view of the camera and for which parallax issues can be neglected. As a consequence, we can assume for any two frames indices $i$ and $j$ in $\lbrace1, \ldots, N\rbrace$ the existence of a true, unknown pairwise transformation $T_{i,j} : \Omega_j \rightarrow \Omega_i$ relating the two images $I_i$ and $I_j$ such that
\begin{equation}
I_j(\textbf{x}) \equiv I_i \circ T_{i,j}(\textbf{x}),
\label{eq:registration_warping}
\end{equation}
where $\equiv$ denotes the pointwise correspondence stating that $I_j(\textbf{x})$ and $I_i \circ T_{i,j}(\textbf{x})$ are two acquisitions of the same point of the imaged scene. Note that, in terms of pixel values, we have $I_i \circ T_{i,j}(\textbf{x}) \neq I_j(\textbf{x})$ in general due to possible imaging noise or illumination changes.
By definition, we have $T_{i,i} = \textrm{Id}$, $T_{i,j}^{-1} = T_{j,i}$ and $T_{i,k} = T_{i,j} \circ T_{j,k}$ for all $i, j, k$ in $\lbrace1, \ldots, N\rbrace$.

After choosing a frame $I_r$ of the sequence as reference, the objective of video mosaicking can be defined as the estimation of the true spatial transformations $\left( T_{r,n}\right)_{1 \leq n \leq N}$ relating each frame $I_n$ of the sequence and the reference frame $I_r$. These $N$ transformations uniquely define the spatial relationship between any two frames $I_i$ and $I_j$ via the identity \begin{equation}
T_{i,j} = T_{r,i}^{-1} \circ T_{r,j}.
\label{eq:global_to_pairwise}
\end{equation}
We denote $\Theta_n$ the transformations $T_{r,n}$ to be estimated. Since $\Theta_r = T_{r,r} = \textrm{Id}$, the mosaicking task reduces to the estimation of the $N-1$ remaining transformations $\left(\Theta_{n}\right)_{n \neq r}$. In this paper, we set the first frame of the sequence as reference, but another reference frame could be used without loss of generality; a strategy to choose an optimal reference frame was, for example, presented in \citet{Xia2017pr}. 

\begin{figure}[t]
\centering
\includegraphics[width=0.58\textwidth]{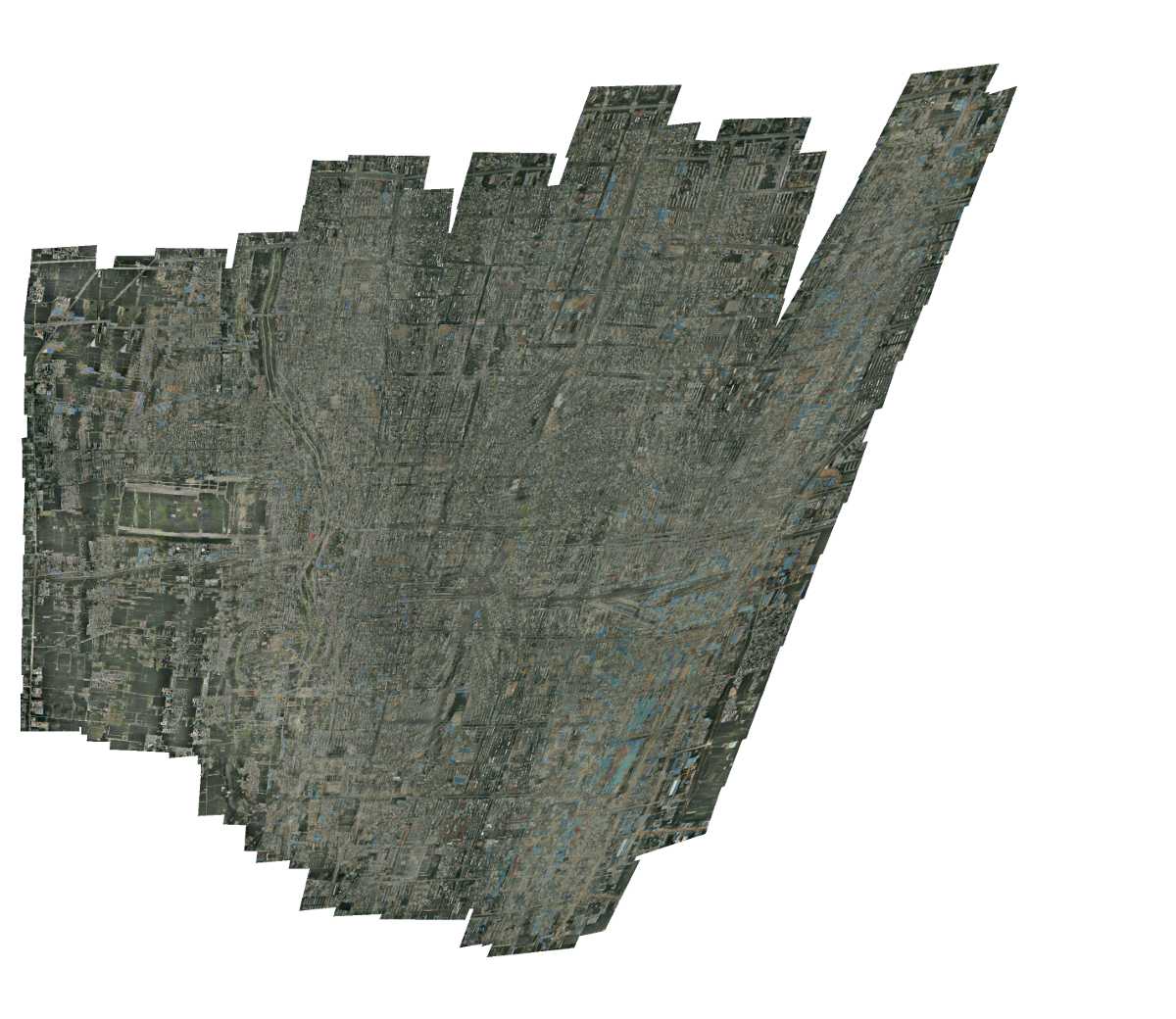}
\caption{\textbf{Sequential mosaicking is prone to drift.} This figure shows a mosaic reconstruction from an aerial imaging sequence of $744$ frames introduced in~\citet{Xia2017pr}, where the trajectory follows a raster scan starting at the bottom left corner and ending at the top right corner. For the displayed reconstruction, only pairs of consecutive frames in the sequence were matched. Although the quality of the alignment is satisfactory on the first frames, minor registration errors between each pair progressively add up, leading to a distorted mosaic. To ensure a more consistent mosaic (as shown in Fig.~\ref{fig:aerial_final_mosaic}), overlapping pairs at a longer range must be registered to further constrain the reconstruction.}
\label{fig:aerial_initial_mosaic}
\end{figure}

In practice, estimating the relative transformation between two frames is done by registering these two frames, which is only successfully feasible if these frames have a sufficient spatial overlap. 
For video mosaicking, one expects the camera motion to be sufficiently slow (in comparison to the acquisition frame rate) to guarantee that at least the consecutive frames $I_{n-1}$ and $I_n$ for $n \in \lbrace 2, \ldots, N \rbrace$ have a sufficient overlap. By registering these consecutive frames, one obtains (noisy) estimates $(\hat{T}_{n-1,n})_{2 \leq n \leq N}$ of the true pairwise transformations $\left(T_{n-1,n}\right)_{2 \leq n \leq N}$. For all $n \in \lbrace{2, \ldots, N\rbrace}$, the global transformation $\Theta_n$ can then be estimated as
\begin{equation}
\hat{\Theta}_n = \hat{T}_{1,2} \circ  \hat{T}_{2,3} \circ \ldots \circ \hat{T}_{n-1,n}.
\label{eq:chain_of_transformations}
\end{equation}
Unfortunately, since the transformations obtained as an output of a registration method are only imperfect estimates of the true pairwise transformations, performing a series of compositions as suggested by Eq.~\ref{eq:chain_of_transformations} leads to an accumulation of errors called \textit{drift}. As a result, the estimated $\hat{T}_n$ may significantly deviate from the true $T_n$ when the number of compositions in Eq.~\ref{eq:chain_of_transformations} is large. Visual drift is especially problematic for the quality of the reconstruction if the trajectory of the camera revisits some parts of the scene at different timepoints, possibly many frames apart in the sequence. In this case, a mosaic built via a composition of registrations displays a lack of global consistency (Fig.~\ref{fig:aerial_initial_mosaic}).

To reduce drift, it is essential to register additional pairs of overlapping frames that are at distant timepoints in the sequence, and to conduct a global optimisation incorporating these additional constraints. Formally, instead of the set $\mathcal{C} = \lbrace(n-1,n)\rbrace_{2 \leq n \leq N}$ made of pairs of consecutive frames, we consider a larger set of pairs $\mathcal{P}$ containing $\mathcal{C}$ for which a pairwise image correspondence is available, i.e. such that $I_i$ and $I_j$ overlap and were registered, leading to an available measured transformation $\hat{T}_{i,j}$. The drift-prone estimation of Eq.~\ref{eq:chain_of_transformations} based on a chain can then be replaced by a globally consistent formulation
\begin{equation}
(\hat{\Theta}_n)_{n \neq r}  = \argmin_{(\Theta_n)_{n \neq r} } \sum_{(i,j) \in \mathcal{P}} d(\Theta_i, \Theta_j, \hat{T}_{i,j}),
\label{eq:general_bundle_adjustment}
\end{equation}
commonly referred to as \textit{bundle adjustment}.
In Eq.~\ref{eq:general_bundle_adjustment}, we assumed given a dissimilarity measure $d$ stating the incompatibility between two global transformations $\Theta_i$ and $\Theta_j$ and a pairwise measurement $\hat{T}_{i,j}$. The measure $d$ can be constructed to take into account the visual content of the images $I_i$ and $I_j$ (e.g., by enforcing the transformation compatibility on a set of extracted keypoints). Intuitively, $d$ must satisfy
\begin{equation}
d(\Theta_i, \Theta_j, \hat{T}_{i,j}) \approx 0 \ \ \text{if} \ \ \hat{T}_{i,j} \approx \Theta_i^{-1} \circ \Theta_j,
\end{equation}
as a consequence of Eq.~\ref{eq:global_to_pairwise}. 
In general, duplicate of pairs could be considered if several registration measurements are available for a given pair of images. In this work, without loss of generality, we consider for simplicity that each pair $(i,j)$ yields at most a single pairwise measurement $\hat{T}_{i,j}$. 


For the minimisation problem of Eq.~\ref{eq:general_bundle_adjustment} to improve over Eq.~\ref{eq:chain_of_transformations} in terms of drift reduction, the set $\mathcal{T} = \lbrace\hat{T}_{i,j}\rbrace_{(i,j) \in \mathcal{P}}$ of available pairwise transformations must contain long-range correspondences, in addition to the estimated transformations between consecutive frames. Establishing an appropriate set of pairwise transformations $\mathcal{T}$ for a given sequence can be challenging: if one only considers the accuracy of the reconstruction, the ideal solution would be to attempt the registration of all pairs of frames in the sequence, and to build the set $\mathcal{T}$ from the results of the successful registrations. However, this strategy raises three key issues:
\begin{description}
\item[\textit{Choice of a registration algorithm}] If the input sequence is visually challenging, it can be difficult to design a registration algorithm which produces reliable estimates $\hat{T}_{i,j}$
for a given pair of long-range overlapping frames. Indeed, in contrast to consecutive frames which are usually of similar appearance by continuity of the trajectory, long-range overlapping frames display a greater variability of illumination and perspective conditions, due to possibly large differences in the pose of the camera when it revisits a certain part of the scene. 
\item[\textit{Reliability assessment of the registration results}]
As a consequence of the difficulty of registering pairs of long-range frames, the accuracy of attempted long-range registrations is not guaranteed. Therefore, a matching algorithm must be coupled with a mechanism able to automatically assess whether the performed registration was successful, in order to prevent the addition of erroneous terms in Eq.~\ref{eq:general_bundle_adjustment}. For example, the output of a pairwise registration algorithm is bound to be wrong if it is run on two frames that do not actually spatially overlap.
\item[\textit{Topology inference}] The number of possible pairwise registrations $\frac{N(N-1)}{2}$ grows quadratically with the number of frames $N$ in the sequence. Therefore, even under the assumption that a reliable pairwise registration algorithm is available, the required computational complexity is very high for long sequences. In fact, an exhaustive registration of all pairs is unnecessary: usually, most frames in the sequence do not overlap and thus cannot be registered in the first place. Inferring the topology of the sequence, i.e. understanding which pairs overlap prior to registration, is crucial to reduce the number of registration attempts and to keep a tractable number of terms in the cost function of Eq.~\ref{eq:general_bundle_adjustment}.
\end{description}
We present in Section~\ref{sec:related_work} a review of the literature on mosaicking with a focus on these three challenges.

\section{Related Work}
\label{sec:related_work}

Mosaicking and image stitching techniques have a long history in computer vision~\citep{szeliski2007image,triggs2000ba}. For some applications, such as in retinal imaging~\citep{richa2014tmi}, the low number of frames does not require the use of a globally-consistent formulation. In these cases, research works focus for example on better visual rendering~\citep{Zhang_2014_CVPR} or on the adaptation to dynamic scenes~\citep{mahe2015motion}. Here, we review more specifically how previous works have addressed the general problem of defining a set of pairwise correspondences $\mathcal{T}$ to be included in a bundle adjustment formulation such as Eq.~\ref{eq:general_bundle_adjustment}. Following the three challenges identified at the end of Section~\ref{sec:image_mosaicking}, we review in Section~\ref{sec:pairwise_registration} the most common strategies for pairwise image registration in mosaicking senarios, and we review in Section~\ref{sec:failure_identification} the methods to assess the reliability of the transformations estimated by these registration algorithms. The existing works on topology inference are discussed in Section~\ref{sec:topology_inference}. A more complete overview of other aspects related to mosaicking can be found in~\citet{szeliski2007image,triggs2000ba}.

\subsection{Pairwise Registration}
\label{sec:pairwise_registration} 

Pairwise registration between images is a crucial building block in mosaicking, and the choice of an appropriate registration method depends on the visual properties of the images. If the contrast conditions allow it, the extraction and matching of salient feature points is one of the most successful strategies~\citep{botterill2010real,Brown2007,Elibol2013ras,Gracias2004ifac,Xia2017pr}. As the images to be registered may overlap only partially in an image stitching context, the use of invariant features is very suitable due to its high robustness to differences in terms of scale or rotation, without the need of an initial estimate of the transformation~\citep{Brown2007}. In more challenging scenarios where salient landmarks cannot be reliably extracted, a direct dense alignment of images can be used instead~\citep{loewke2011vivo,lovegrove2010real,molina2014persistent,Peter2018ijcars,richa2014tmi,shum1998construction}, for example via a least-square minimisation on the pixel intensities~\citep{lovegrove2010real,richa2014tmi} or on the image gradient orientations~\citep{Peter2018ijcars}.
More recently, learning-based registration has also been investigated in a mosaicking context to increase the robustness to the challenging visual conditions occuring during fetal surgery~\citep{bano2020deep}.

\subsection{Reliability Assessment}
\label{sec:failure_identification}

In practice, an automatic registration algorithm remains prone to occasional failures, which prompts the introduction of a mechanism identifying and rejecting such failures. In the seminal work of \citet{Brown2007}, the registration is performed by matching keypoints using a RANSAC criterion which differentiates correct matches from erroneous matches and naturally provides a rejection mechanism for feature-based registration~\citep{Atasoy2008,botterill2010real,garcia2016icra}. However, such a mechanism is not readily available for registration strategies based on dense alignment, in which case previous works mostly resort to heuristics on the cost function~\citep{Gracias2004ifac,loewke2011vivo,Peter2018ijcars,Sawhney1998eccv} or to application-driven strategies such as the detection of the centerline in retinal imaging~\citep{can2002pami,yang2004cvpr}. The direct estimation of registration uncertainty given two registered images~\citep{kybic2009bootstrap}, which can be seen as a generalisation of reliability assessment, is also a growing direction of research, especially in the context of medical image registration~\citep{muenzing2012supervised,risholm2013bayesian,sokooti2016accuracy}.

If allowed by the application, bringing a human in the loop can be an effective option to guarantee the validity of a registration. This is for example feasible if the mosaic can be created offline without strong time constraints, or if one desires to build a database of annotated correspondences. Via human supervision, the output of an automated registration can be visually verified and, if needed, corrected by interactively annotating visual correspondences between the two frames. In this context, an interactive approach to annotate correspondences between two frames was proposed by \citet{jegelka2014interactive}. Their approach is complementary to ours, in the sense that we focus on the interactive retrieval of pairs of frames to be annotated and on the underlying issues in terms of informativeness measure and uncertainty modelling, rather than on registration. Once a pair is suggested and shown to the agent, an interactive approach as proposed by Jegelka et al. could be used in our framework as an annotation tool to match the two suggested frames.


\subsection{Topology Inference}
\label{sec:topology_inference}

The problem of topology inference plays a crucial role as soon as the number of images to be stitched is large enough to make the accumulation of registration errors problematic. The importance of moving away from sequential mosaicking was identified in the seminal work of \citet{Sawhney1998eccv}, in which the term \textit{topology inference} was introduced. To reconstruct a mosaic on a sequence of 184 frames, \citet{Sawhney1998eccv} identified both the need to perform a global optimisation instead of a sequential alignment, and the need to keep the number of long-range correspondences low for computational efficiency. An iterative approach was proposed which alternates between the addition of soft constraints and the resolution of bundle adjustment.
After dense alignment, a threshold on the loss function is then used as heuristic to accept or discard the performed registration. 
Following this seminal work, other approaches used the current reconstruction as an indicator of the topology of the sequence~\citep{can2002pami,kang2000graph,loewke2011vivo,Marzotto2004cvpr,prokopetc2016reducing}.  Since the cost of bundle adjustment updates between iterations can be problematic, \citet{Gracias2004ifac} proposed an affine bundle adjustment formulation allowing online updates of the bundle adjustment results directly on the mosaic. Alternatively, efficient mosaic updates can be obtained based on the separating axis theorem~\citep{Kekec2014ras,tella2019pruning}. \citet{lovegrove2010real} also achieved real-time efficiency using parallelisation of local and global alignment. The complexity of bundle adjustment updates can also be achieved by restricting it to a subset of key frames in the sequence~\citep{steedly2005efficiently}. If salient features can be extracted and tracked along the sequence, visual SLAM approaches and image-to-mosaic mesh updates~\citep{civera2009drift,davison2007monoslam,kim2006real}
have also been considered to maintain in real time an informative model of the observed scene.

The aforementioned approaches rely, at each iteration, on the currently reconstructed mosaic to predict whether two frames overlap. Therefore, such approaches cannot handle arbitrarily long loops in the sequence for which drift cannot be avoided until a location is revisited for the first time. To overcome this issue, uncertainty on the probability of overlap can be taken into account. To the best of our knowledge, only \citet{Elibol2013ras}  proposed such a probabilistic overlap model to query pairs of frames to be matched. However, this probabilistic model was then used to avoid the suggestion of false positives by suggesting pairs with very high chances of overlap which, intrinsically, does not allow the suggestion of informative long-range pairs for which the positional uncertainty is higher.

As positional information is, by nature, mostly insufficient to efficiently identify long-range overlapping pairs, a popular strategy consists in incorporating another source of overlap information that is independent of the performed registrations and of the current mosaic. Depending on the application, this additional overlap predictor can take several forms, such as an external position sensor placed on the camera~\citep{tella2018probabilistic,tella2019pruning}, some prior knowledge on the trajectory of the camera~\citep{civera2009drift,mahe2013viterbi,molina2014persistent}, or an appearance model linking the probability of frame overlap to their visual similarity~\citep{botterill2010real,garcia2016icra,Ho2007,Peter2018ijcars,Xia2017pr}.

\begin{figure*}[t]
\centering
\includegraphics[width=0.95\textwidth]{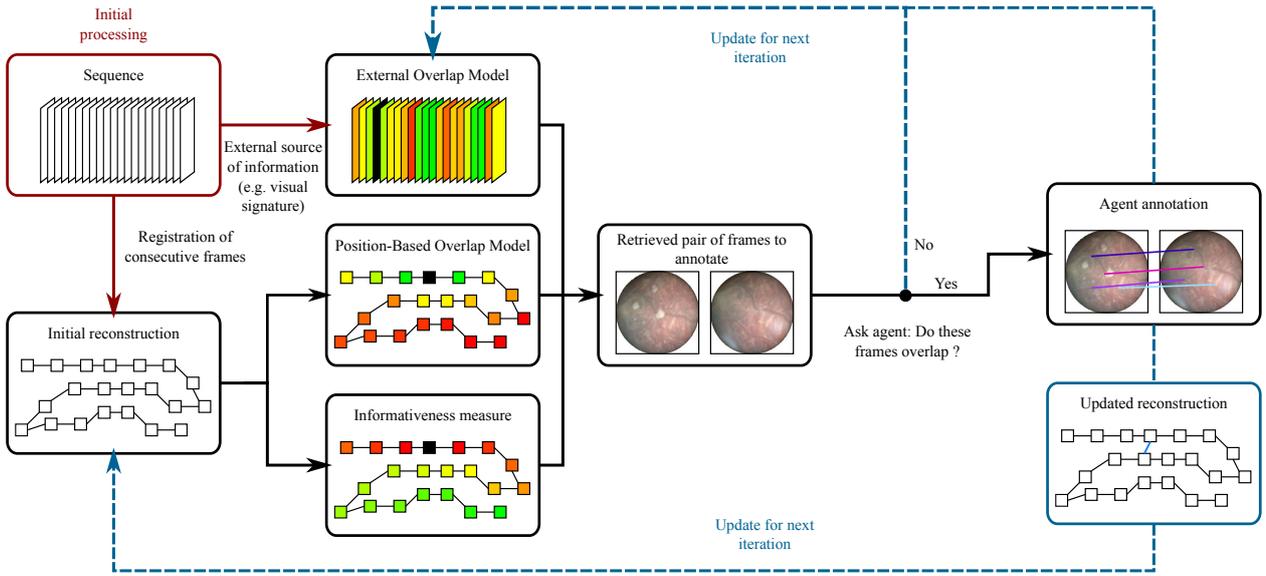}
\caption{\textbf{Overview of our interactive framework}.  A mosaic is initally reconstructed by registering consecutive frames of the sequence and an external model of frame overlap is formed based on a registration-independent source of information (such as an image appearance model). At each iteration, a position-based overlap model and an informativeness score are inferred from the current reconstruction. By combining the two complementary models of frame overlap and the informativeness measure, the pair of frames yielding the highest expected annotation reward is suggested for annotation. The agent states whether the two frames indeed overlap, and this information is used to update the external overlap model. If the frames overlap, the agent additionally registers the retrieved frames by providing landmark correspondences. The reconstructed mosaic is then updated by including these new pairwise correspondences, before proceeding with the next iteration.}
\label{fig:abstract}
\end{figure*}

\section{Contributions}
\label{sec:contributions}

In this paper, we address a scenario for the active annotation of pairwise correspondences in a sequence, where an annotating oracle agent (e.g. a human user) provides reliable annotations at a high annotation cost. This novel scenario has implications in terms of requirements for pairwise registration and topology inference. By the oracle nature of the agent, the long-range correspondences furnished by the agent are considered highly reliable, i.e. subject to an annotation noise of small, known variance. However, the cost of the pairwise annotations made by the agent imposes stronger constraints on the topology inference aspect. 
To keep the number of agent queries as low as possible, it is thus of high importance to carefully select each queried pair by avoiding redundant suggestions and making use as much as possible of the reliable information provided by the agent on each suggested pair.

To summarise, we have, on one hand, access to reliable annotations of pointwise correspondences (or of the absence thereof) for any given pair of images. However, on the other hand, this reliable information is associated to an annotation cost and can only be queried sparsely. This leads us to introduce the following contributions:
\begin{enumerate}

\item We propose a probabilistic model predicting whether two frames of the sequence spatially overlap, which is crucial to \textit{only suggest annotable pairs of frames} to the agent. Building on the two main ideas proposed in the literature to infer the topology of a sequence (Section~\ref{sec:topology_inference}), this model combines two complementary types of information: the current belief on the position of each frame on the mosaic, and an external source of overlap information defined as the visual similarity between images prior to registration.

\item We define an annotation reward stating how informative the annotation of each pair would be at a given iteration. This reward is designed as to \textit{choose the most relevant pair to annotate}, in order to avoid the query of annotations that are redundant with previously acquired ones. Moreover, our chosen reward is designed as to satisfy an asymptotic tradeoff ensuring mathematical stability as the number of frames and the resulting positional uncertainty on the mosaic increases.

\item We introduce update strategies which revise on-the-fly the two complementary models of frame overlap. These updates allow to \textit{leverage the reliable information provided by the agent} at each iteration. In particular, if a suggested pair was reported as not overlapping and thus not annotable, this information is incorporated to prevent the repetition of similar unannotable queries.
\end{enumerate}

\section{Methods}
\label{sec:methods}

\subsection{Problem Statement}
\label{sec:problem_statement}


As described in Section~\ref{sec:image_mosaicking}, we consider a sequence $(I_n)_{1 \leq n \leq N}$ and a set $\mathcal{T}_1 = \lbrace\hat{T}_{i,j}\rbrace_{(i,j) \in \mathcal{P}_1}$ of initial estimated transformations between some pairs of frames that are known to overlap, and such that each frame of the sequence is represented in $\mathcal{P}_1$. For example, $\mathcal{P}_1$ can be the set of consecutive pairs $\mathcal{C} = \lbrace(n-1,n)\rbrace_{2 \leq n \leq N}$, but can also be extended to registrations with multiple neighbours.

Starting from this initial set of registrations, our approach iteratively suggests pairs of frames whose geometrical correspondence, or absence thereof, is to be annotated by an agent. At iteration $k \geq 1$, we denote $\mathcal{P}_k$ the set of pairs which have been annotated, either by the agent or during the initial registration phase. This set is partitioned as $\mathcal{P}_k = \mathcal{P}^+_k \cup \mathcal{P}^-_k$. $\mathcal{P}^+_k$ is the set pairs for which a spatial overlap and visual correspondence are known, i.e. the pairs initially registered and the pairs annotated as overlapping by the agent. On the contrary, $\mathcal{P}^-_k$ is the set of queries for which no actual visual overlap was found by the agent. Initially,  $\mathcal{P}_1^+ = \mathcal{P}_1$ and $\mathcal{P}_1^- = \emptyset$. We denote $\mathcal{T}_k = \lbrace\hat{T}_{i,j}\rbrace_{(i,j) \in \mathcal{P}_k^+}$ the current set of estimated spatial transformations which are available for the known overlapping pairs.

At iteration $k \geq 1$, the strategy suggesting the next pair of frames to be annotated runs as follows (Fig.~\ref{fig:abstract}):
\begin{enumerate}
\item Given the set of pairwise transformations $\mathcal{T}_k$, the estimated global transformations $(\hat{\Theta}_n)_{n \neq r}$ are computed via bundle adjustment (Eq.~\ref{eq:general_bundle_adjustment}), together with an uncertainty estimate of these global positions in the reconstruction (Section~\ref{sec:bundle_adjustment}).
\item Denoting $O_{ij}$ the probabilistic event ``$I_i$ and $I_j$ spatially overlap'', we compute for each pair $(i,j)$ two different probabilities of overlap based on two complementary sources of information:
\begin{itemize}
\item a \textit{position-based} probability $P^{\textrm{pos}}(O_{ij} \mid \mathcal{T}_k)$ which predicts whether $I_i$ and $I_j$ overlap based on the topology of the current reconstructed mosaic and on the reconstruction uncertainty, which are both computed from the available correspondences $\mathcal{T}_k$ (Section~\ref{sec:position_based_overlap_probability}),
\item an independent overlap model $P^{\textrm{ext}}(O_{ij} \mid I_i, I_j, \bold{w}_k)$ of parameters $\bold{w}_k$ based on an \textit{external} source of information, irrespective of registration.  In this work, we compute a registration-independent measure of visual similarity between the images $I_i$ and $I_j$, combining the respective visual signatures of $I_i$ and $I_j$ defined via a weighted bag-of-words model of learned parameters $\bold{w}_k$ (see Section~\ref{sec:content_based_overlap_probability}). 
\end{itemize}
Considering these two sources of overlap information as independent, we obtain a combined probability of overlap between any two images $I_i$ and $I_j$, given as
\begin{equation}
P(O_{ij} \mid I_i, I_j, \bold{w}_k, \mathcal{T}_k) = \\
 P^{\textrm{ext}}(O_{ij} \mid I_i, I_j, \bold{w}_k) P^{\textrm{pos}}(O_{ij} \mid \mathcal{T}_k).
\label{eq:proba_overlap}
\end{equation}
Note that this overlap model is made dependent of the iteration $k$ via the parameters $\bold{w}_k$ and the available correspondences $\mathcal{T}_k$. As the agent annotates additional pairs, the overlap probabilities are updated to incorporate this newly acquired information.
\item We compute for each possible pair $(i,j)$ an uncertainty measure $U_{ij}(\mathcal{T}_k)$ stating how informative the annotation of the frames $I_i$ and $I_j$ would be at the current iteration $k$, for a given set of available correspondences $\mathcal{T}_k$. Intuitively, the higher the position uncertainty between $I_i$ and $I_j$, the more informative the acquisition of a pairwise correspondence between these two frames would be, and thus the higher $U_{ij}(\mathcal{T}_k)$ is (Section~\ref{sec:annotation_reward}).
\item Equipped with this uncertainty measure, we would like to suggest the pair with the highest uncertainty. However, by nature, only overlapping pairs are annotable by the agent. Querying a pair of frames which do not overlap is thus a ``wasted'' iteration in terms of agent resources. To encode this aspect, we define the annotation reward $R^{(k)}_{ij}$ as $0$ if the frames do not overlap, and are therefore not annotable, and as $U_{ij}(\mathcal{T}_k)$ if they are annotable. We then query annotations from the agent on the pair $(i_k,j_k)$ showing the highest expected reward, namely
\begin{equation}
(i_k,j_k) = \argmax_{(i,j)} \mathbb{E}[R^{(k)}_{ij}],
\label{eq:pair_highest_reward}
\end{equation}
where, following from our assumptions,
\begin{equation}
\mathbb{E}[R^{(k)}_{ij}] = P(O_{ij} \mid I_i, I_j, \bold{w}_k, \mathcal{T}_k) U_{ij}(\mathcal{T}_k).
\label{eq:expected_reward}
\end{equation}
\item The agent signals whether the images $I_{i_k}$ and $I_{j_k}$ forming the suggested pair overlap. If they do, the agent annotates the pair by providing landmark correspondences on the two frames which yields a reliable and accurate estimate $\hat{T}_{(i_k,j_k)}$ of the transformation relating these two overlapping frames. This transformation is then added to $\mathcal{T}_k$, i.e. $\mathcal{T}_{k+1} = \mathcal{T}_k \cup \lbrace\hat{T}_{(i_k,j_k)}\rbrace$, $\mathcal{P}^+_{k+1} = \mathcal{P}^+_k \cup \lbrace(i_k,j_k)\rbrace$ and $\mathcal{P}^-_{k+1} = \mathcal{P}^-_k$. If the suggested pair of frames is declared as not annotable by the agent, no additional correspondence can be added. In this case, we define $\mathcal{T}_{k+1} = \mathcal{T}_k$, $\mathcal{P}^+_{k+1} = \mathcal{P}^+_k$ and $\mathcal{P}^-_{k+1} = \mathcal{P}^-_k \cup \lbrace(i_k,j_k)\rbrace$.
\item Based on the agent feedback, the parameters $\bold{w}_k$ of the external overlap model are reestimated. Intuitively, if the suggested pairs did not overlap, the overlap model is revised to avoid similar erroneous suggestions in the future. In other words, the predictive model stating the probability of overlap is learned online based on the information provided by the agent (Section~\ref{sec:update_external_model}). 
\end{enumerate}
The process consisting of the steps (1-6) above is repeated iteratively until an annotation budget from the agent is exceeded, or until a certain uncertainty threshold on the reconstructed mosaic is reached. 
The following subsections expose in details the different components of our method forming the suggestion strategy.

\subsection{Transformation Model}
\label{sec:transformation_model}

The general considerations on mosaicking presented in Section~\ref{sec:image_mosaicking} rely on the choice and parametrisation of a transformation model. 
Under a planarity assumption, two images of the scene are related by a homography with 8 degrees of freedom, which is therefore a suitable transformation model in theory and used in some applications~\citep{botterill2010real,Marzotto2004cvpr,tella2018probabilistic}. However, in practice, other transformation spaces with less degrees of freedom are often used instead. In increasing order of complexity, examples of such alternative transformation models include translations~\citep{mahe2013viterbi}, rigid transformations~\citep{richa2014tmi}, similarity transformations~\citep{Elibol2013ras,garcia2016icra,molina2014persistent} and affine transformations~\citep{choe2006icpr,Gracias2004ifac,Peter2018ijcars,prokopetc2016reducing}. Occasionally, transformations with more than 8 degrees of freedom can be used to account for application-specific non-planarity effects, such as in retinal imaging~\citep{can2002pami,yang2004cvpr}. Opting for transformations with a restricted number of degrees of freedom is especially appropriate in the case of sequences with a large number of frames, as these transformations are naturally less prone to drift due to their lower complexity~\citep{Xia2017pr}. Moreover, as the primary aim of mosaicking is to provide a 2D image with a larger field-of-view than the camera and not a reconstructed 3D object, the presence of distortion effects can be acceptable for practical purposes as long as the topology of the scene is preserved~\citep{Peter2018ijcars}.


In this work, similar to \citet{choe2006icpr,Gracias2004ifac,Peter2018ijcars,prokopetc2016reducing}, we use affine transformations. This choice is primarily motivated by the fact that affine bundle adjustment can be formulated as a linear least-square problem and that, as a result, new pairwise correspondences can be efficiently added via a recursive linear least-squares formulation~\citep{Gracias2004ifac}. In an interactive setting like ours, this computational advantage is crucial to be able to efficiently update the bundle adjustment estimates between two iterations and ensure a short waiting time to the annotating agent. 


An affine transformation $T$ is defined by $6$ parameters $(t_1, \ldots, t_6) \in \mathbb{R}^6$ and maps by definition an input point $\textbf{x} = (x,y)^\mathrm{T} \in \mathbb{R}^2$ to an output $T(\textbf{x}) \in \mathbb{R}^2$ given by $T(\textbf{x}) = (t_1 x + t_2 y + t_3 , t_4 x + t_5 y + t_6)^\mathrm{T}$. This definition can be rewritten in matrix form by expressing $\textbf{x}$ and $T(\textbf{x})$ in homogeneous coordinates as $\tilde{\textbf{x}} = (x,y,1)^\mathrm{T}$ and $\tilde{T}(\textbf{x}) = (T_x(\textbf{x}),T_y(\textbf{x}),1)^\mathrm{T}$, such that $\tilde{T}(\textbf{x}) = \bold{T} \tilde{\textbf{x}}$ where
\begin{equation}
\bold{T} =
\begin{pmatrix}
t_1 & t_2 & t_3 \\
t_4 & t_5 & t_6 \\
0 & 0 & 1
\end{pmatrix}.
\label{eq:def_affine_homogeneous}
\end{equation}
In particular, this representation allows the composition and inversion of affine transformations through matrix multiplication and inversion, respectively.

\subsection{Notations}
\label{sec:notations}

Before describing our methods in details in the next sections, we introduce here some additional notations used in this paper. Bold lowercase symbols denote column vectors and bold uppercase symbols denote matrices. We identify an affine transformation $T$ with the corresponding $3 \times 3$ matrix $\bold{T}$. For a positive integer $d$, $\bold{0}_{d}$ and $\bold{1}_{d}$ are the column vectors of size $d$ whose coordinates are all equal to $0$ and $1$ respectively, and $\bold{I}_d$ is the identity matrix of size $d \times d$. Given two integers $i$ and $j$, $\delta_{ij}$ is the Kronecker delta defined as $\delta_{ij} = 1$ if $i = j$ and $\delta_{ij} = 0$ otherwise. $\bar{\delta}_{ij} = 1 - \delta_{ij}$ is its complement, and we also define $\delta_{i > j} = 1$ if $i > j$ and $\delta_{i > j} = 0$ otherwise. For $1 \leq j \leq d$, we denote $\bold{e}_{j}^{(d)}$ the $j$-th column vector of the standard basis of $\mathbb{R}^d$, i.e a vector of size $d$ whose $j$-th coordinate is $\delta_{jd}$.  Similarly $\bold{E}^{(n \times p)}_{ij}$ is the $n \times p$ matrix whose entry $(i,j$) is $1$ and $0$ everywhere else. We denote
\begin{equation}
\bold{P}_{c \leftarrow h} = \begin{pmatrix}
1 & 0 & 0 \\
0 & 1 & 0 \\
\end{pmatrix}
\end{equation}
the matrix which converts homogeneous coordinates into Cartesian coordinates, i.e. such that $\bold{x} = \bold{P}_{c \leftarrow h} \tilde{\bold{x}}$ for any $\bold{x} \in \mathbb{R}^2$ represented as $\tilde{\bold{x}} \in \mathbb{R}^3$ in homogeneous coordinates. Given a $3 \times 3$ affine matrix $\bold{T}$ defined as in Eq.~\ref{eq:def_affine_homogeneous}, we define
\begin{equation}
\bold{T}^{(\textrm{lin})} = 
\begin{pmatrix}
t_1 & t_2 \\
t_4 & t_5
\end{pmatrix} = \bold{P}_{c \leftarrow h} \bold{T} \bold{P}_{c \leftarrow h}^{\mathrm{T}}
\end{equation}
the linear part of $\bold{T}$.
We denote $\vecrow$ the operator which performs a rowwise vectorisation of a matrix, such that for example
\begin{equation}
\vecrow(\bold{T}) = (t_1, t_2, t_3, t_4, t_5, t_6, 0,0,1)^{\mathrm{T}}.
\end{equation}
We then have, for any matrices $\bold{A}$, $\bold{X}$ and $\bold{B}$ such that $\bold{A}\bold{X}\bold{B}$ exists, the identity
\begin{equation}\vecrow(\bold{A}\bold{X}\bold{B}) = (\bold{A} \otimes \bold{B}^\mathrm{T}) \vecrow(\bold{X}),
\label{eq:vectorise_product}
\end{equation}
where $\otimes$ denotes the Kronecker product between two matrices.
For affine transformations more specifically, we define the restricted vectorisation operator
\begin{equation}
\vecaff(\bold{T}) = (t_1, \ldots, t_6)^\mathrm{T} = \vecrow(\bold{P}_{c \leftarrow h} \bold{T})
\end{equation}
furnishing the vector of coefficients parametrising $\bold{T}$.
It is also easy to show that $\vecaff$ and $\vecrow$ are related by
\begin{equation}
\vecrow(\bold{T}) =  (\bold{P}_{c \leftarrow h}^{\mathrm{T}}\otimes \bold{I}_3) \vecaff(\bold{T}) + \bold{e}_{9}^{(9)}.
\label{eq:vec_to_vec_6}
\end{equation}
More generally, we extend the operators $\vecrow$ and $\vecaff$ to an ordered collection $(\bold{T}_n)_{1 \leq n \leq N}$ of $N$ affine transformations by stacking vertically the $N$ corresponding vectors. For example, $\vecaff\left[ (\bold{T}_n)_{1 \leq n \leq N}\right]$ is the column vector of size $6N$ obtained by stacking the $N$ vectors  $\vecaff(\bold{T}_n)$ of size $6$ each.  With this definition, we have 
\begin{equation}
\vecaff(\bold{T}_n) = \left(\bold{e}_{n}^{(N)} \otimes \bold{I}_6\right)^{\mathrm{T}} \vecaff\left[ (\bold{T}_n)_{1 \leq n \leq N}\right].
\label{eq:vec_6_to_vec_6_collection}
\end{equation}
In the case of two transformations $\bold{T}_1$ and $\bold{T}_2$ only, the vector $\vecaff\left[ (\bold{T}_n)_{1 \leq n \leq 2}\right]$ is more simply written  $\vecaff(\bold{T}_1, \bold{T}_2)$.


%

\subsection{Bundle Adjustment and Uncertainty Estimates}
\label{sec:bundle_adjustment}

This section describes our adopted strategy for the general problem of reconstructing a mosaic from a set of pairwise correspondences, i.e. our particular instantiation of the general formalism recalled in Section~\ref{sec:image_mosaicking}. We give closed-form solutions for solving affine bundle adjustment based on pairwise matches, which reformulates the formalism of \citet{Gracias2004ifac} by expliciting the interplay between each provided match and the parameters of the global transformations encoding the reconstruction. Based on these results, we additionally derive a novel closed-form probabilistic model of the predicted position of each frame on the canvas based on covariance propagation. This uncertainty model can be seen as an extension of the model proposed by \citet{Elibol2013ras} for similarity transformations and with a different bundle adjustment formulation. The probabilistic framework described in this section is the basis of our position-based overlap probability and our measure of pairwise informativeness, respectively presented in Section~\ref{sec:position_based_overlap_probability} and Section~\ref{sec:annotation_reward}.

\subsubsection{Pairwise Correspondences}
\label{sec:affine_pairwise_registration}

\begin{figure}[t]
\begin{center}
\def\svgwidth{1\linewidth}
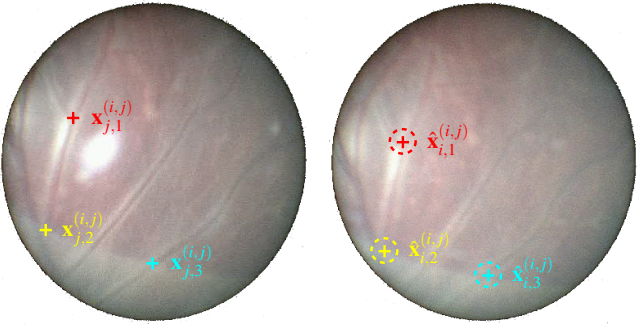
\end{center}
\caption{\textbf{Pairwise correspondences and uncertainty model}. Two images $I_j$ (left) and $I_i$ (right) are spatially registered by a human annotator (or another oracle agent) providing a set of pairwise correspondences. In this example taken from our fetoscopy dataset, $3$ correspondences are manually annotated. The uncertainty on the provided matches is modeled by an isotropic Gaussian distribution on each provided landmark in $I_i$.}
\label{fig:correspondences}
\end{figure}

A building block of mosaicking is the estimation $\hat{T}_{i,j}$ of the true transformation $T_{i,j}$ relating two overlapping images $I_i$ and $I_j$. To account for both keypoint-based and dense alignment in a unified formulation and be able to incorporate uncertainty estimates, we do not explicitly model $\hat{T}_{i,j}$ and, instead, reason in terms of pairwise correspondences between the two images $I_i$ and $I_j$ defined as follows. For a given image pair, a pairwise correspondence is established via a set of $L_{ij}$ locations $\mathcal{X}^{(i,j)}_j = \lbrace\textbf{x}^{(i,j)}_{j,1}, \cdots, \textbf{x}^{(i,j)}_{j,L_{ij}}\rbrace$ in the input image $I_j$ and their corresponding estimated matching locations $\hat{\mathcal{X}}^{(i,j)}_i = \lbrace\hat{\textbf{x}}^{(i,j)}_{i,1}, \cdots, \hat{\textbf{x}}^{(i,j)}_{i,L_{ij}}\rbrace$ in the image $I_i$ (Fig.~\ref{fig:correspondences}).
This formulation encompasses three special cases of interest: (\textit{i})~feature-based registration, where $\mathcal{X}^{(i,j)}_j$ is the set of landmarks composing inlier matches, (\textit{ii})~registration via dense alignment, where $\mathcal{X}^{(i,j)}_j$ is a possibly sparse predefined grid of control points, and (\textit{iii})~human annotations, where $\mathcal{X}^{(i,j)}_j$ and $\hat{\mathcal{X}}^{(i,j)}_i$ are manually provided by the user. 

To model the uncertainty on a given set of pairwise correspondences, we proceed as follows~\citep{kanatani2004uncertainty,raguram2009exploiting}. Considering the set of extracted landmarks in $I_j$ as fixed, we assume that the set $\hat{\mathcal{X}}^{(i,j)}_i$ of the estimated locations in $I_i$ differs from the set of true matches $\mathcal{X}^{(i,j)}_i$ via an isotropic, i.i.d Gaussian noise, namely
\begin{equation}
\hat{\textbf{x}}^{(i,j)}_{i,l} \sim \mathcal{N}\left( \textbf{x}^{(i,j)}_{i,l} , \sigma^2 \bold{I}_2\right) 
\label{eq:pointwise_uncertainty}
\end{equation}
for all $l \in \left\lbrace 1, \ldots, L_{ij} \right\rbrace$.
As both the agent annotations and the initial registrations between consecutive frames are considered to be reliable, the value of $\sigma$ is assumed small and identical for all pairs, typically of the order of the pixel. In this paper, we set $\sigma = 1$ for simplicity. However, our framework is compatible with variances that are image-specific or even landmark-specific if such an additional information is available.

%
%

\subsubsection{Affine Bundle Adjustment}
\label{sec:affine_bundle_adjustment}

The bundle adjustment formulation introduced in Eq.~\ref{eq:general_bundle_adjustment} requires the definition of a measure of compatibility between transformations. If $\textbf{x}^{(i,j)}_{i,l}$ and $\textbf{x}^{(i,j)}_{j,l}$ are two locations which visually correspond, by definition we have $\textbf{x}^{(i,j)}_{i,l} = T_{i,j}(\textbf{x}^{(i,j)}_{j,l})$ where $T_{i,j}$ is the true transformation relating $I_i$ and $I_j$. Using Eq.~\ref{eq:global_to_pairwise}, this equality can be rewritten
\begin{equation}
T_{r,i}(\textbf{x}^{(i,j)}_{i,l}) = T_{r,j}(\textbf{x}^{(i,j)}_{j,l}).
\label{eq:pairwise_correspondence}
\end{equation}
From Eq.~\ref{eq:pairwise_correspondence} and under our affine transformation model, we consider the following bundle adjustment formulation based on all the pairwise correspondences available for a given sequence as originally used by \citet{Gracias2004ifac}:
\begin{equation}
(\hat{\boldsymbol{\Theta}}_n)_{n \neq r} = \argmin_{(\boldsymbol{\Theta}_n)_{n \neq r}} \sum_{(i,j) \in \mathcal{P}_k^+} \sum_{l=1}^{L_{ij}} \Vert \boldsymbol{\Theta}_j \tilde{\textbf{x}}^{(i,j)}_{j,l} - \boldsymbol{\Theta}_i  \tilde{\hat{\textbf{x}}}^{(i,j)}_{i,l} \Vert^2,
\label{eq:affine_bundle_adjustment}
\end{equation}
where we remind that $\boldsymbol{\Theta}_r = \bold{I}_{3}$ by definition of the reference frame, and that $\tilde{\bold{x}} \in \mathbb{R}^3$ corresponds to the expression of $\bold{x}$ in homogeneous coordinates (Section~\ref{sec:transformation_model}). 

Equation~\ref{eq:affine_bundle_adjustment} can be rewritten and solved as a linear least-square minimisation problem as follows. We denote $\tilde{\bold{X}}^{(i,j)}_i$ the $3 \times L_{ij}$ matrix $ \left(\tilde{\hat{\textbf{x}}}^{(i,j)}_{i,1} \ldots \tilde{\hat{\textbf{x}}}^{(i,j)}_{i,L_{ij}} \right)$. We similarly define the $3 \times L_{ij}$ matrix $\tilde{\bold{X}}^{(i,j)}_j$ and the $2 \times L_{ij}$ matrices $\hat{\bold{X}}^{(i,j)}_i$ and  $\bold{X}^{(i,j)}_j$. With these definitions, Eq.~\ref{eq:affine_bundle_adjustment} can be rewritten as an optimisation problem over the $6(N-1)$ parameters of the affine transformations as
\begin{align}
\vecaff[(\hat{\boldsymbol{\Theta}}_n)_{n \neq r}] &= \argmin_{\bm{\theta} \in \mathbb{R}^{6(N-1)}} \sum_{(i,j) \in \mathcal{P}_k^+} \Vert \bold{A}_{ij}^{\mathrm{T}} \bm{\theta} - \bold{b}_{ij} \Vert^2 \label{eq:bundle_adjustment_affine_vectorised_first_line}\\
&= \argmin_{\bm{\theta} \in \mathbb{R}^{6(N-1)}} \left( \frac{1}{2} \bm{\theta}^{\mathrm{T}} \bold{S} \bm{\theta} - \bold{v}^{\mathrm{T}} \bm{\theta}\right) ,
\label{eq:bundle_adjustment_affine_vectorised}
\end{align}
where
\begin{equation}
\bold{A}_{ij} = \bar{\delta}_{jr} \bold{e}_{j - \delta_{j>r}}^{(N-1)} \otimes  \bold{I}_2 \otimes \tilde{\bold{X}}^{(i,j)}_j 
 - \bar{\delta}_{ir} \bold{e}_{i - \delta_{i>r}}^{(N-1)} \otimes  \bold{I}_2 \otimes \tilde{\hat{\bold{X}}}^{(i,j)}_i,
\label{eq:definition_of_A}
\end{equation}
\begin{equation}
\bold{b}_{ij} =  \delta_{ir} \vecrow\left(\hat{\bold{X}}^{(i,j)}_i\right) - \delta_{jr} \vecrow\left(\bold{X}^{(i,j)}_j\right),
\label{eq:definition_of_b}
\end{equation}
\begin{equation}
\bold{S} = \sum_{(i,j) \in \mathcal{P}_k^+}  \bold{A}_{ij} \bold{A}_{ij}^{\mathrm{T}}
\label{eq:definition_of_S}
\end{equation}
and
\begin{equation}
\bold{v} = \sum_{(i,j) \in \mathcal{P}_k^+} \bold{A}_{ij} \bold{b}_{ij}.
\label{eq:definition_of_v}
\end{equation}
The equivalence between Eq.~\ref{eq:affine_bundle_adjustment} and its vectorised form Eq.~\ref{eq:bundle_adjustment_affine_vectorised_first_line} is not straightforward. We refer to Appendix~\ref{appendix:bundle_adjustment_affine_vectorised} for the full derivations. The equivalence between Eq.~\ref{eq:bundle_adjustment_affine_vectorised_first_line} and Eq.~\ref{eq:bundle_adjustment_affine_vectorised} is easily shown by expanding the squared norms in Eq.~\ref{eq:bundle_adjustment_affine_vectorised}:
\begin{equation}
\Vert \bold{A}_{ij}^{\mathrm{T}} \bm{\theta} - \bold{b}_{ij} \Vert^2 = \left( \bold{A}_{ij}^{\mathrm{T}} \bm{\theta} - \bold{b}_{ij}\right)^\mathrm{T} \left(  \bold{A}_{ij}^{\mathrm{T}} \bm{\theta} - \bold{b}_{ij}\right).
\end{equation}
Setting the gradient of the objective function to $\bold{0}$, solving Eq.~\ref{eq:bundle_adjustment_affine_vectorised} amounts to solving $\bold{S}\bm{\theta} = \bold{v}$, which can be done efficiently using the Cholesky decomposition of $\bold{S}$. The form of Eq.~\ref{eq:definition_of_S} and \ref{eq:definition_of_v} also reveals that the addition of a pairwise measurement consists of a low-rank update of $\bold{S}$ and $\bold{v}$ (since $\rank(\bold{A}_{ij}) = 6$). It follows that the Cholesky decomposition of $\bold{S}$ used to solve Eq.~\ref{eq:bundle_adjustment_affine_vectorised}, and consequently the results of the bundle adjustment, can be efficiently updated between iterations.

In comparison to \citet{Gracias2004ifac}, our proposed derivations above have the advantage of making explicit the individual contribution of each provided pointwise correspondence $(\hat{\textbf{x}}^{(i,j)}_{i,l},\textbf{x}^{(i,j)}_{j,l})$ in the resulting reconstruction defined by the $6(N-1)$ parameters $\vecaff[(\hat{\boldsymbol{\Theta}}_n)_{n \neq r}]$. We exploit these expressions to estimate the uncertainty on the reconstruction, as described in the following section.

\subsubsection{Uncertainty Estimation}

We propagate the uncertainty model on the correspondences (Eq.~\ref{eq:pointwise_uncertainty}) to obtain an uncertainty on the bundle adjustment estimates as follows. For a given pair $(i,j) \in \mathcal{P}_k^+$, we concatenate  all the landmark locations $\hat{\textbf{x}}^{(i,j)}_{i,1}, \ldots, \hat{\textbf{x}}^{(i,j)}_{i,L_{ij}}$ into a single vector $\hat{\bold{m}}_i^{(i,j)}$. These measurement vectors, one for each image pair, are themselves concatenated to form a single measurement vector $\hat{\bold{m}}$. We similarly denote $\bold{m}$ the vector formed by all the unknown true positions $\textbf{x}^{(i,j)}$. Assuming independence between the pairwise matches, Eq.~\ref{eq:pointwise_uncertainty} can be immediately translated to the vector $\bold{m}$ as
\begin{equation}
\bold{m} \sim \mathcal{N}\left(\hat{\bold{m}}, \boldsymbol{\Sigma}_\bold{m}\right),
\end{equation}
where
\begin{equation}
\boldsymbol{\Sigma}_\bold{m} = 
\diag_{(i,j) \in \mathcal{P}_k^+} \left[ \sigma^2 \bold{I}_{2 L_{ij}}\right].
\label{eq:Sigma_m}
\end{equation}
With these notations, the bundle adjustment formulation of Eq.~\ref{eq:bundle_adjustment_affine_vectorised} can be rewritten
\begin{equation}
\vecaff[(\hat{\boldsymbol{\Theta}}_n)_{n \neq r}] = \argmin_{\bm{\theta} \in \mathbb{R}^{6(N-1)}}  F(\bm{\theta},\hat{\bold{m}}),
\end{equation}
with
\begin{equation}
F(\bm{\theta},\hat{\bold{m}}) = \frac{1}{2}\bm{\theta}^{\mathrm{T}} \bold{S}(\hat{\bold{m}}) \bm{\theta} - \bold{v}(\hat{\bold{m}})^{\mathrm{T}} \bm{\theta},
\label{eq:bundle_adjustment_cost_function}
\end{equation} 
where we explicited the dependency of $\bold{S}$ and $\bold{v}$ in the measured correspondences.
Similarly to previous works on mosaicking~\citep{Elibol2013ras}, we follow Haralick's method for covariance propagation~\citep{haralick1994} which consists in a first-order linearisation of the gradient of the objective function $F(\bm{\theta},\bold{m})$
around its minimum. 
We obtain
\begin{equation}
\vecaff[(\boldsymbol{\Theta}_n)_{n \neq r}]  \sim \mathcal{N}\left(  \vecaff[(\hat{\boldsymbol{\Theta}}_n)_{n \neq r}] , \boldsymbol{\Sigma}_{(\boldsymbol{\Theta}_n)_{n \neq r}} \right)
\label{eq:proba_ba_estimates}
\end{equation}
where
\begin{align}
\boldsymbol{\Sigma}_{(\boldsymbol{\Theta}_n)_{n \neq r}} &= \bold{S}^{-1} \bold{F}_{\bm{\theta} \bold{m}} \boldsymbol{\Sigma}_\bold{m} \bold{F}_{\bm{\theta} \bold{m}}^\mathrm{T} \bold{S}^{-1} \\
&= \sigma^2 \bold{S}^{-1} \bold{F}_{\bm{\theta} \bold{m}} \bold{F}_{\bm{\theta} \bold{m}}^\mathrm{T} \bold{S}^{-1}.
\label{eq:covariance_propagation_haralick}
\end{align}
$\bold{F}_{\bm{\theta} \bold{m}}$ is the matrix of the cross-derivatives with respect to $\bm{\theta}$ and $\bold{m}$ as defined in~\citet{haralick1994} and obtained as follows: denoting $M = |\mathcal{P}_k^+|$ the number of input measurements, $\bold{F}_{\bm{\theta} \bold{m}}$ is the concatenation of $M$ column blocks $\bold{F}^{(i,j)}_{\bm{\theta} \bold{m}}$ of size $6(N-1) \times 2 L_{ij}$, where the pairs $(i,j)$ are ordered in the same fashion as in $\boldsymbol{\Sigma}_\bold{m}$. For $l \in \lbrace 1, \ldots, L_{ij}\rbrace$, $d \in \lbrace1, 2\rbrace$ and $c = 2(l-1)+d$, the $c^{\textrm{th}}$ column of $\bold{F}^{(i,j)}_{\bm{\theta} \bold{m}}$ is given by
\begin{multline}
(\bold{F}^{(i,j)}_{\bm{\theta} \bold{m}})_{c} = \left[\frac{\partial \bold{A}_{i j}}{\partial x_c} \bold{A}^{\mathrm{T}}_{i j} + \bold{A}_{i j} \frac{\partial \bold{A}^{\mathrm{T}}_{i j}}{\partial x_c}\right] \vecaff[(\hat{\boldsymbol{\Theta}}_n)_{n \neq r}] \\
- \frac{\partial\bold{A}_{i j}}{\partial x_c} \bold{b}_{ij} - \bold{A}_{i j} \frac{\partial\bold{b}_{i j}}{\partial x_c},
\label{eq:F_uv}
\end{multline}
where we denote $\frac{\partial}{\partial x_c}$ the derivative with respect to the entry of coordinates $(d,l)$ in $\hat{\bold{X}}^{(i,j)}_i$. By linearity, we have
\begin{equation}
\frac{\partial \bold{A}_{i j}}{\partial x_c} 
= - \bar{\delta}_{ir} \bold{e}_{i - \delta_{i>r}}^{(N-1)} \otimes  \bold{I}_2 \otimes \bold{E}^{(3 \times L_{ij})}_{d,l}
\end{equation}
and
\begin{align}
\frac{\partial \bold{b}_{i j}}{\partial x_c} = \delta_{ir} \vecrow\left(\bold{E}^{(2 \times L_{ij})}_{d,l}\right) = \delta_{ir} \bold{e}^{(2 L_{ij})}_{(d-1)L_{ij} + l}.
\end{align}
Note that these derivatives are sparse matrices so that each column can be exactly and quickly computed via Eq.~\ref{eq:F_uv} with only a few operations. 



\subsection{Position-Based Overlap Probability}
\label{sec:position_based_overlap_probability}


In this section, we introduce a model of frame overlap providing, for any pair of frames $(I_i,I_j)$, a probability $P^{\textrm{pos}}(O_{ij} \mid \mathcal{T})$ that $I_i$ and $I_j$ overlap based on the available pairwise correspondences $\mathcal{T}$. This probability makes use of the probabilistic information on the position of each frame on the canvas derived in the previous section and defined by the mean $\vecaff[(\hat{\boldsymbol{\Theta}}_n)_{n \neq r}]$ and covariance matrix $\boldsymbol{\Sigma}_{(\boldsymbol{\Theta}_n)_{n \neq r}}$, following Eq.~\ref{eq:proba_ba_estimates}.

In our context, the notion of ``overlap'' must be understood as ``sufficient overlap as to be annotable by the agent''. We define this notion as follows: we say that $I_i$ sufficiently overlaps with $I_j$ when the centre of $I_i$ possesses a corresponding point in $I_j$. Equivalently, using $I_j$ as reference frame, this amounts to saying that the centre of $I_i$ in this reference frame belongs to the image domain $\Omega$. Denoting $\boldsymbol{\gamma}_{ij} = (\gamma_{ij}^{(x)},\gamma_{ij}^{(y)})^\mathrm{T}$ the 2D coordinates of the centre of $I_i$ mapped in the reference frame defined by $I_j$, we have thus defined
\begin{equation}
P^{\textrm{pos}}(O_{ij} \mid \mathcal{T}) = P(\boldsymbol{\gamma}_{ij} \in \Omega \mid \vecaff[(\hat{\boldsymbol{\Theta}}_n)_{n \neq r}], \boldsymbol{\Sigma}_{(\boldsymbol{\Theta}_n)_{n \neq r}}) .
\label{eq:definition_position_probability}
\end{equation}
From Eq.~\ref{eq:registration_warping}, $\boldsymbol{\gamma}_{ij}$ is defined as $T_{ij}(\boldsymbol{\gamma}_{ij}) = \boldsymbol{\gamma}$, where $\boldsymbol{\gamma}$ is the centre of the reference domain $\Omega$, i.e. the centre of $I_j$ if $I_j$ acts as a reference frame. Since $T_{ij}^{-1} = T_{ji}$, we have $\boldsymbol{\gamma}_{ij} = T_{ji}(\boldsymbol{\gamma})$. Expressing this equality in homogeneous coordinates yields
\begin{align}
\boldsymbol{\gamma}_{ij} &= \bold{P}_{c \leftarrow h} \bold{T}_{j,i} \tilde{\boldsymbol{\gamma}} \\
&= \bold{P}_{c \leftarrow h} \bold{T}_{j,r} \bold{T}_{i,r}^{-1} \tilde{\boldsymbol{\gamma}} \\
&= \bold{P}_{c \leftarrow h} \boldsymbol{\Theta}_{j}^{-1} \boldsymbol{\Theta}_i \tilde{\boldsymbol{\gamma}}.
\label{eq:definition_gamma_ij}
\end{align}
The objective of this section is to derive, based on the knowledge on the distribution of $\boldsymbol{\Theta}_i$ and $\boldsymbol{\Theta}_j$ described by Eq.~\ref{eq:proba_ba_estimates}, the probability $P(\boldsymbol{\gamma}_{ij} \in \Omega)$ defined in Eq.~\ref{eq:definition_position_probability}. We first describe how to compute an estimate of this probability in Section~\ref{sec:position_overlap_probability_montecarlo}. In Section~\ref{sec:lower_upper_bounds}, we derive analytical approximate lower and upper bounds of this probability, which is useful both for computational efficiency and to obtain mathematical insights on its behaviour and accordingly define our informativeness measure (see Section~\ref{sec:annotation_reward}).

\begin{figure*}[t]
\centering
\subfloat[Analytical lower bound]{\includegraphics[width=0.32\textwidth]{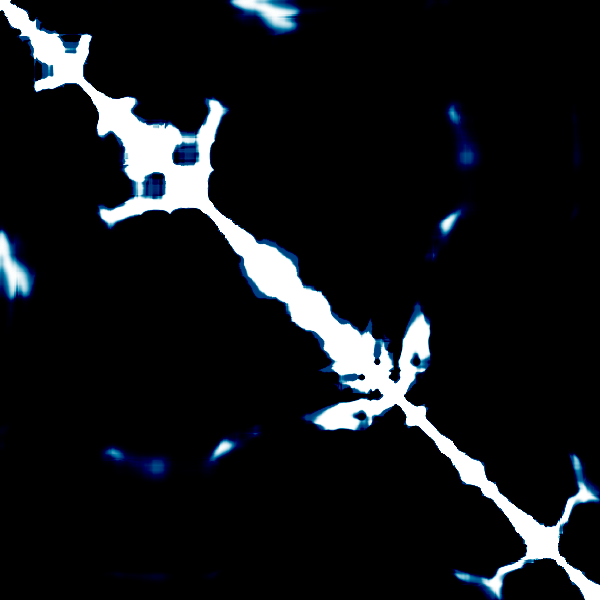}}\hfill
\subfloat[Sampling-based probability of overlap]{\includegraphics[width=0.32\textwidth]{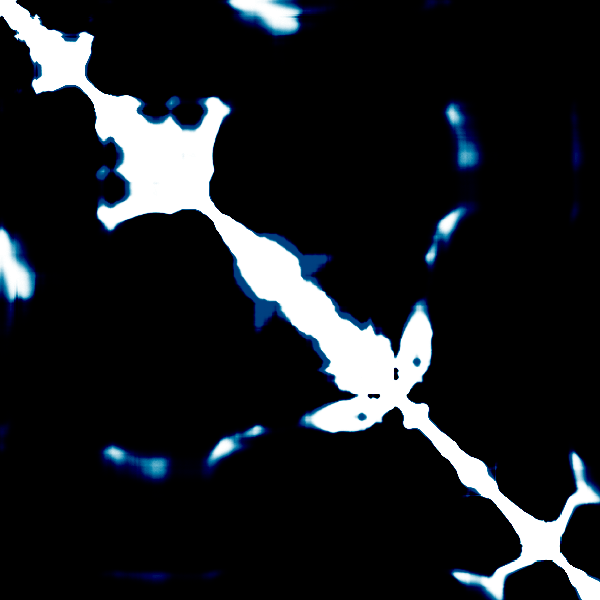}}\hfill
\subfloat[Analytical upper bound]{\includegraphics[width=0.32\textwidth]{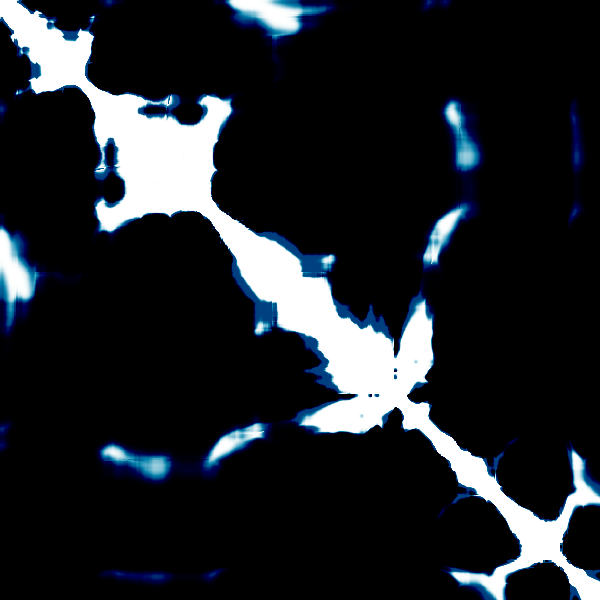}}\hfill
\caption{\textbf{Position-based overlap probability and approximate closed-form bounds.} Matrix representation of the probability of frame overlap. Each entry $(i,j)$ of (b) states the probability of overlap of the frames $I_i$ and $I_j$ based on the current mosaic reconstruction and its uncertainty model, obtained by sampling (Section~\ref{sec:position_overlap_probability_montecarlo}). (a) and (c) show lower and upper bounds of these probabilities (Section~\ref{sec:lower_upper_bounds}).}
\label{fig:position_overlap_models}
\end{figure*}

\subsubsection{Numerical Approximation}
\label{sec:position_overlap_probability_montecarlo}
As a consequence of the multivariate normal distribution of Eq.~\ref{eq:proba_ba_estimates}, marginalising over the position of the remaining frames trivially gives
\begin{equation}
\vecaff(\boldsymbol{\Theta}_i, \boldsymbol{\Theta}_j)
 \sim  \mathcal{N} \left(
  \vecaff(\hat{\boldsymbol{\Theta}}_i, \hat{\boldsymbol{\Theta}}_j )
 , \boldsymbol{\Sigma}_{({\boldsymbol{\Theta}}_i, \boldsymbol{\Theta}_j)} \right),
\label{eq:marginalise_pair}
\end{equation}
where $\boldsymbol{\Sigma}_{({\boldsymbol{\Theta}}_i, \boldsymbol{\Theta}_j)}$ is the $12 \times 12$ submatrix of $\boldsymbol{\Sigma}_{(\boldsymbol{\Theta}_n)}$ obtained by retaining the $6 \times 6$ covariance blocks of $\boldsymbol{\Sigma}_{(\boldsymbol{\Theta}_n)}$ corresponding to $\boldsymbol{\Theta}_i$ and $\boldsymbol{\Theta}_j$.
To the best of our knowledge, under a Gaussian assumption on $\vecaff(\boldsymbol{\Theta}_i, \boldsymbol{\Theta}_j)$, there is no closed-form expression of the probability $P \left( \boldsymbol{\gamma}_{ij} \in \Omega\right)$ given by Eq.~\ref{eq:definition_gamma_ij} in the general case. Instead, a Monte Carlo approach could be followed to numerically estimate $P \left( \boldsymbol{\gamma}_{ij} \in \Omega\right)$ by sampling a series of transformations $\boldsymbol{\Theta}_i^{(s)}$ and $\boldsymbol{\Theta}_j^{(s)}$ according to Eq.~\ref{eq:marginalise_pair}, and test numerically for each of these samples whether they verify $\bold{P}_{c \leftarrow h} (\boldsymbol{\Theta}_j^{(s)})^{-1} \boldsymbol{\Theta}_i^{(s)}  \tilde{\boldsymbol{\gamma}} \in \Omega$. This numerical strategy has the advantage to give an estimate as accurate as desired by increasing the number of samples.

Although the suggested Monte Carlo strategy would give an accurate numerical estimation of the probability of Eq.~\ref{eq:definition_position_probability}, this computation remains too expensive to be done for all pairs of frames as is necessary to evaluate the overlap probability of each pair. In the interest of computational efficiency, we perform instead a first-order approximation of $\boldsymbol{\gamma}_{ij}$ with respect to the parameters $\vecaff(\boldsymbol{\Theta}_i, \boldsymbol{\Theta}_j)$, where the non-linearity originates from the inversion of $\boldsymbol{\Theta}_j$ (see Eq.~\ref{eq:definition_gamma_ij}). After linearisation, lower and upper bounds of the linearised probability can be computed yielding computational advantages over the Monte Carlo approach (Sec~\ref{sec:lower_upper_bounds}). The linear approximation of $\boldsymbol{\gamma}_{ij}$ is given by
\begin{equation}
\boldsymbol{\gamma}_{ij} \approx \hat{\boldsymbol{\gamma}}_{ij} + \bold{J}_{ij} \left[ 
  \vecaff(\boldsymbol{\Theta}_i, \boldsymbol{\Theta}_j) - \vecaff(\hat{\boldsymbol{\Theta}}_i, \hat{\boldsymbol{\Theta}}_j)\right],
\label{eq:linear_approximation_centre_distribution}
\end{equation}
where 
\begin{equation}
\hat{\boldsymbol{\gamma}}_{ij}  = \bold{P}_{c \leftarrow h} \hat{\boldsymbol{\Theta}}_j^{-1} \hat{\boldsymbol{\Theta}}_i \tilde{\boldsymbol{\gamma}}
\end{equation}
and
$\bold{J}_{ij}$ is the $2 \times 12$ Jacobian matrix evaluated at $\vecaff(\hat{\boldsymbol{\Theta}}_i, \hat{\boldsymbol{\Theta}}_j)$. $\bold{J}_{ij}$ is the columnwise concatenation of the two $2 \times 6$ blocks $\bold{J}^{(i)}_{ij}$ and $\bold{J}^{(j)}_{ij}$ defined as
\begin{equation} 
\bold{J}^{(i)}_{ij} = (\hat{\boldsymbol{\Theta}}_j^{\textrm{(lin)}})^{-1} \otimes  \tilde{\boldsymbol{\gamma}}^{\mathrm{T}}
\label{eq:jacobian_gamma_i}
\end{equation}
and 
\begin{equation}
\bold{J}^{(j)}_{ij} = - (\hat{\boldsymbol{\Theta}}_j^{\textrm{(lin)}})^{-1} \otimes  \tilde{\hat{\boldsymbol{\gamma}}}_{ij}^{\mathrm{T}}.
\label{eq:jacobian_gamma_j}
\end{equation}
The derivations leading to the expressions of $\bold{J}^{(i)}_{ij}$ and $\bold{J}^{(j)}_{ij}$ can be found in Appendix~\ref{appendix:jacobian}.
Since $\vecaff(\boldsymbol{\Theta}_i, \boldsymbol{\Theta}_j)$ follows a $12$-dimensional Gaussian distribution (Eq.~\ref{eq:marginalise_pair}), it follows from Eq.~\ref{eq:linear_approximation_centre_distribution} that, under our linear approximation, $\boldsymbol{\gamma}_{ij}$ follows a bivariate Gaussian distribution, namely
\begin{equation}
\boldsymbol{\gamma}_{ij} \sim \mathcal{N}\left( \hat{\boldsymbol{\gamma}}_{ij}, \boldsymbol{\Sigma}_{\boldsymbol{\gamma}_{ij}} \right) 
\label{eq:centre_distribution}
\end{equation}
where
\begin{equation}
\boldsymbol{\Sigma}_{\boldsymbol{\gamma}_{ij}} = \bold{J}_{ij} \boldsymbol{\Sigma}_{({\boldsymbol{\Theta}}_i, \boldsymbol{\Theta}_j)} \bold{J}_{ij}^{\mathrm{T}}.
\label{eq:centre_covariance_matrix}
\end{equation}
From now on, we identify $P(\boldsymbol{\gamma}_{ij} \in \Omega)$ with the probability that the linearised version of $\boldsymbol{\gamma}_{ij}$ given by Eq.~\ref{eq:centre_distribution} belongs to $\Omega$.
Based on Eq.~\ref{eq:centre_distribution}, the computation of $P \left( \boldsymbol{\gamma}_{ij} \in \Omega\right)$ reduces to the integration of a bivariate Gaussian over an arbitrary  (off-centered and non-eigenvector-aligned) rectangle $\Omega$ (Fig.~\ref{fig:notations_overlap}). Again, this is not feasible in closed form to the best of our knowledge, and we follow instead a numerical, two dimensional Monte Carlo strategy combined with analytical lower and upper bounds derived below.

\subsubsection{Analytical Lower and Upper Bounds}
\label{sec:lower_upper_bounds}

Even in a two-dimensional setting, a Monte Carlo approximation cannot be done for all pairs of the sequence without leading to a considerable waiting time between iterations. In this section we derive approximate lower and upper bounds of $P( \boldsymbol{\gamma}_{ij} \in \Omega)$. The advantage of having access to these bounds is twofold:
\begin{enumerate}
\item An upper bound on the overlap probability provides, at suggestion time, an upper bound on the expected reward (Eq.~\ref{eq:expected_reward}). As a result, the computationally costlier sampling step can be avoided if we know in advance that the expected reward cannot exceed the expected reward of the highest-ranked pair found so far. By dynamically maintaining in a heap structure the best pairs that were found, we can obtain at low computational cost the top $K$ pairs for any predefined number $K \geq 1$.
\item  These bounds provide insights into the mathematical behaviour of the overlap probability, which helps us in return to choose an appropriate informativeness score for each pair (Section~\ref{sec:annotation_reward}).
\end{enumerate} 
\begin{figure}[t]
\begin{center}
\def\svgwidth{1\linewidth}
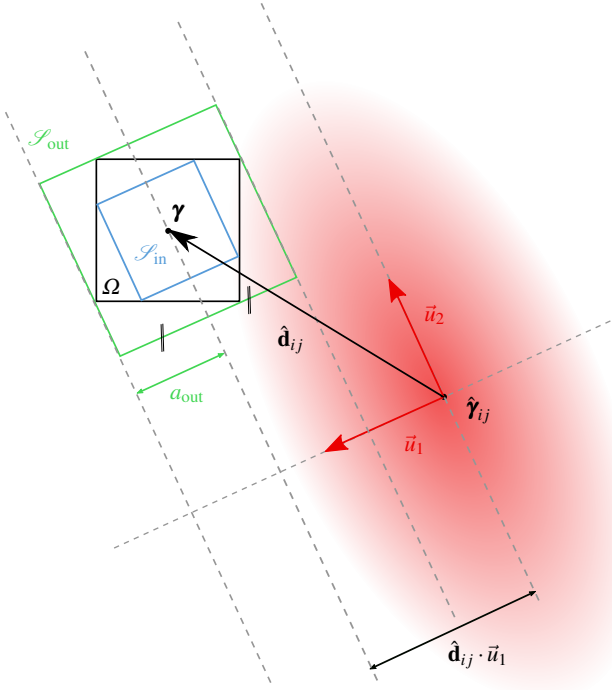
\end{center}
\caption{The centre $\boldsymbol{\gamma}_{ij}$ of the image $I_i$, taken in the reference frame of $I_j$ centered on $\boldsymbol{\gamma}$, follows (after linearisation) a bivariate Gaussian distribution of mean position $\hat{\boldsymbol{\gamma}}_{ij}$. $\vec{u}_1$ and $\vec{u}_2$ denote the eigenvectors of the corresponding covariance matrix, and $\nu_1$ and $\nu_2$ are the associated eigenvalues. Although computing $P(\boldsymbol{\gamma}_{ij} \in \Omega)$ in closed form is not feasible (to the best of our knowledge), closed form bounds can be obtained by considering two squares $\mathcal{S}_{\textrm{in}}$ and $\mathcal{S}_{\textrm{out}}$ whose sides are aligned with $\vec{u}_1$ and $\vec{u}_2$. Thereby, the bivariate Gaussian distribution governing $\boldsymbol{\gamma}_{ij}$ splits into two independent Gaussian variables, allowing to use the expression of the cumulative distribution of univariate Gaussian distributions based on the error function $\erf$, as shown in Eq.~\ref{eq:position_overlap_product}.}
\label{fig:notations_overlap}
\end{figure}
The derivation of the bounds on $P(\boldsymbol{\gamma}_{ij} \in \Omega)$ can be done as follows.
The real-valued, symmetric positive definite covariance matrix $\boldsymbol{\Sigma}_{\boldsymbol{\gamma}_{ij}}$ has two orthogonal eigenvectors $\vec{u}_1$ and $\vec{u}_2$ and positive eigenvalues $\nu_1$ and $\nu_2$, defining two independent Gaussian distributions along $\vec{u}_1$ and $\vec{u}_2$. We denote $\hat{\bold{d}}_{ij} = \boldsymbol{\gamma} - \hat{\boldsymbol{\gamma}}_{ij}$ the displacement vector between the mean positions (in the reference frame of $I_j$) of the centres of $I_j$ and $I_i$, respectively. Without loss of generality, we choose the orientation of $\vec{u}_1$ and $\vec{u}_2$ such that $\hat{\bold{d}}_{ij}\cdot \vec{u}_1 \geq 0$ and $\hat{\bold{d}}_{ij} \cdot \vec{u}_2 \geq 0$.
We consider two squares $\mathcal{S}_{\textrm{in}}$ and $\mathcal{S}_{\textrm{out}}$, respectively contained in $\Omega$ and containing $\Omega$, whose sides are aligned with $\vec{u}_1$ and $\vec{u}_2$ (see Fig.~\ref{fig:notations_overlap}). By construction, since $\mathcal{S}_{\textrm{in}} \subseteq \Omega \subseteq \mathcal{S}_{\textrm{out}}$, we have
\begin{equation}
P(\boldsymbol{\gamma}_{ij} \in \mathcal{S}_{\textrm{in}}) \leq P(\boldsymbol{\gamma}_{ij} \in \Omega) \leq P(\boldsymbol{\gamma}_{ij} \in \mathcal{S}_{\textrm{out}}).
\label{eq:bounds_generic}
\end{equation}
Moreover, for any square $\mathcal{S}$ whose sides are aligned with the eigenvectors of $\boldsymbol{\Sigma}_{\boldsymbol{\gamma}_{ij}}$, $P(\boldsymbol{\gamma}_{ij} \in \mathcal{S})$ can be expressed by splitting the two independent Gaussians governing the coordinates of $\boldsymbol{\gamma}_{ij}$ in the coordinate system centered on $\boldsymbol{\gamma}$ and aligned with $\vec{u}_1$ and $\vec{u}_2$. If we denote $a$ the half-length of the sides of $\mathcal{S}$, we have
\begin{equation}
P(\boldsymbol{\gamma}_{ij} \in \mathcal{S}) 
 = \prod_{k=1}^2 P(|Z_k - \hat{\bold{d}}_{ij} \cdot \vec{u}_k| \leq a ),
\label{eq:position_overlap_product_dot}
\end{equation}
where $Z_k \sim \mathcal{N}(0,\sqrt{\nu_k})$.  The cumulative distribution function of the $Z_k$ can be expressed with the error function $\erf$ so that Eq.~\ref{eq:position_overlap_product_dot} can be rewritten
\begin{equation}
P(\boldsymbol{\gamma}_{ij} \in \mathcal{S}) 
= \prod_{k=1}^2 f_a(\hat{\bold{d}}_{ij}  \cdot \vec{u}_k,\nu_k)
\label{eq:position_overlap_product}
\end{equation}
where
\begin{equation}
f_a(x,\nu) = \frac{1}{2} \left[ \erf\left( \frac{x + a}{\sqrt{2 \nu}}\right)  - \erf\left( \frac{x - a}{\sqrt{2 \nu}}\right) \right].
\label{eq:function_f_overlap}
\end{equation}
Applying Eq.~\ref{eq:position_overlap_product} to the chosen squares $\mathcal{S}_{\textrm{in}}$ and $\mathcal{S}_{\textrm{out}}$ and denoting $a_{\textrm{in}}$ and $a_{\textrm{out}}$ their respective half-lengths, Eq.~\ref{eq:bounds_generic}
finally gives the desired lower and upper bounds on our approximate position-based probability of frame overlap, i.e.
\begin{equation}
\Lambda_{ij}(\mathcal{T}) \leq  P^{\textrm{pos}}(O_{ij} \mid \mathcal{T}) \leq \Upsilon_{ij}(\mathcal{T}),
\label{eq:bounds_position_probability}
\end{equation}
where
\begin{equation}
\Lambda_{ij}(\mathcal{T}) = \prod_{k=1}^2 f_{a_{\textrm{in}}}(\hat{\bold{d}}_{ij} \cdot \vec{u}_k,\nu_k)
\label{eq:lower_bound_position_probability}
\end{equation}
and
\begin{equation}
\Upsilon_{ij}(\mathcal{T}) = \prod_{k=1}^2 f_{a_{\textrm{out}}}(\hat{\bold{d}}_{ij} \cdot \vec{u}_k,\nu_k).
\label{eq:upper_bound_position_probability}
\end{equation}

\subsection{External Overlap Probability}
\label{sec:content_based_overlap_probability}

The position-based overlap probability introduced in the previous subsection predicts the reliability of the position of each frame on the canvas based on the collected pairwise registrations. Naturally, as a consequence of drift, such a probability decreases as the number of compositions required to infer the relative position of $I_i$ and $I_j$ increases. For arbitrarily large sequences, the position-based probability model alone is thus not sufficient, as it would initially lack informativeness on temporally distant images. As discussed in Section~\ref{sec:topology_inference}, a common approach towards the retrieval of long-range correspondences is to incorporate a second, complementary model of overlap which is based on another source of information than the performed pairwise registrations. We refer to this additional model as \textit{external}.

\subsubsection{External Model Based on Visual Similarity}

\begin{figure}[t]
\centering
\includegraphics[width=0.45\textwidth]{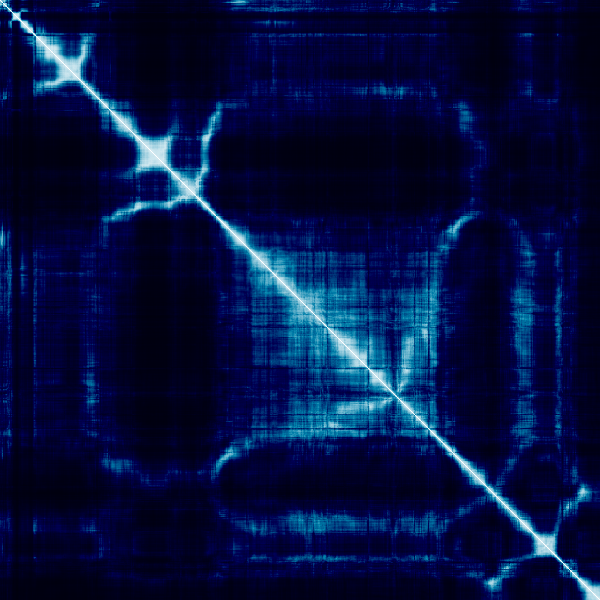}
\caption{\textbf{Overlap probability based on visual similarity.} Each entry $(i,j)$ displays the external overlap probability of overlap for the frames $I_i$ and $I_j$ of our fetoscopy sequence, based on their visual similarity.}
\label{fig:bow_fetoscopy}
\end{figure}

Our interactive setup is a priori compatible with various kinds of external information, e.g. the position and orientation of the camera, if available. In this work, we follow one of the most common approaches in the literature~\citep{Elibol2013ras,garcia2016icra,Ho2007,Peter2018ijcars,Xia2017pr} 
and model this external probability of frame overlap based on the visual similarity between frames, which we encode using a bag of words descriptor. However, unlike in the aforementioned approaches, we desire to keep this model flexible and online trainable, in order to leverage the reliable agent feedback to correct possible inaccuracies of the model, due to challenging or ambiguous visual appearance within the sequence. To do so, we use a learned, weighted bag of words model inspired by the Pairwise Constrained Component Analysis (PCCA) introduced by \citet{mignon2012cvpr}. We consider that a representation based on bags of visual words is available and that a visual dictionary of size $D$ was computed offline (e.g. on the sequence of interest), such that the visual content of each image $I_n$ can be represented by a $D$-dimensional signature $\bold{s}^{(n)} \in \mathbb{R}_+^D $. We do not make any particular assumption on the methodology used to extract these signatures, which can be either application-dependent or learned with a deep learning model~\citep{zagoruyko2015learning}. In our case, we use a generic learned descriptor using convex optimisation introduced by \citet{Simonyan} and readily available in OpenCV~\citep{opencv_library}, which showed to be equally effective for our two datasets in spite of their visual differences. We assume that the signatures $\bold{s}^{(n)}$ are normalised such that $\Vert \bold{s}^{(n)}\Vert_2 = 1$. Given a vector of weights $\bold{w}$ of size $D$, the measure of weighted dissimilarity $\Delta_{\bold{w}}$ between $I_i$ and $I_j$ is then defined as
\begin{align}
\Delta_{\bold{w}}(I_i,I_j) &= \sqrt{\left( \bold{s}^{(i)} - \bold{s}^{(j)} \right)^{\mathrm{T}} \diag(\bold{w})  \left( \bold{s}^{(i)} - \bold{s}^{(j)} \right)} \\
&= \sqrt{\sum_{d = 1}^D w_d \left( s^{(i)}_d - s^{(j)}_d\right)^2}.
\label{eq:bow_distance}
\end{align}
In other words, the vector $\bold{w}$ assigns a relevance weight to each word in the created visual dictionary, i.e. each coordinate of the image signatures, and $\Delta_{\bold{w}}(I_i,I_j)$ is thus a learned metric of similarity between images, with the parameters $\bold{w}$ being the learned model parameters. As in~\citet{mignon2012cvpr}, we convert the distance $\Delta_{\bold{w}}(I_i,I_j)$ into a probability of visual similarity, i.e. a probability of overlap in our context, via the application of a generalised logistic function $h_{\beta}$. This leads to
\begin{align}
P^{\textrm{ext}}(O_{ij} \mid I_i, I_j, \bold{w}) &= h_{\beta}\left(1 - \Delta_{\bold{w}}^2(I_i,I_j) \right) \\
&= h_{\beta}\left( 1 - \bold{w}^{\mathrm{T}} \boldsymbol{\delta}_2(\bold{s}^{(i)},\bold{s}^{(j)})\right),
\label{eq:bow_probability}
\end{align}
where
$\boldsymbol{\delta}_2(\bold{s}^{(i)},\bold{s}^{(j)})$ is the column vector of size $D$ of $d^{\textrm{th}}$ coordinate equal to  $(s^{(i)}_d - s^{(j)}_d)^2$, and where
\begin{equation}
h_{\beta}(x) = \frac{1}{1 + e^{- \beta x}}
\label{eq:generalised_logistic_function}
\end{equation}
for all $x \in \mathbb{R}$. An example of visual similarity matrix obtained through this model is shown in Fig.~\ref{fig:bow_fetoscopy}.

\subsubsection{Model Update}
\label{sec:update_external_model}

Our initial visual similarity weights $\bold{w}_{1}$ are set to $\bold{1}_D$. With this uniform weighting, the quantity $1 - D_{\bold{w}}^2(I_i,I_j)$ reduces (up to a constant) to the cosine similarity measure used in the literature~\citep{Ho2007,Peter2018ijcars}. At each iteration $k$, we revise the weights of our similarity measure by minimising a log-loss which encourages the external model to match the overlap information provided by the agent so far. More precisely, the weight parameters for the next iteration are set to
\begin{equation}
\bold{w}_{k+1} =  \argmin_{\bold{w} \geq \bold{w}_1} \ l_k(\bold{w}),
\label{eq:objective_function}
\end{equation}
where
\begin{equation}
l_k(\bold{w}) = - \sum_{(i,j) \in \mathcal{P}_k^+} \log P_{ij}(\bold{w}) - \sum_{(i,j) \in \mathcal{P}_k^-} \log\left(1 - P_{ij}(\bold{w})  \right)
\label{eq:log_loss_1}
\end{equation}
and $P_{ij}(\bold{w}) = h_{\beta}\left( 1 - \bold{w}^{\mathrm{T}} \boldsymbol{\delta}(\bold{s}_i, \bold{s}_j)\right) $ is the external probability of overlap as defined in Eq.~\ref{eq:bow_probability}. The minimisation of the loss $l_k$ encourages the parameters $\bold{w}$ of the external model to yield a low probability of overlap on the set of image pairs $\mathcal{P}_k^-$ that were annotated as non-overlapping by the agent, and a high probability of overlap on the set of known overlapping pairs $\mathcal{P}_k^+$ (either annotated by the agent or consecutive frames of the sequence). Noting that
\begin{equation}
1 - P_{ij}(\bold{w}) = h_{\beta}\left(\bold{w}^{\mathrm{T}} \boldsymbol{\delta}(\bold{s}_i, \bold{s}_j) - 1 \right)
\end{equation}
and denoting $o_{ij} \in \lbrace-1,1\rbrace$ a label stating the overlap of a pair $(i,j)$, the cost function can be compactly rewritten as
\begin{equation}
l_k(\bold{w}) = \sum_{(i,j) \in \mathcal{P}_k} \log \left(1 + \exp\left[ \beta o_{ij} (\bold{w}^{\mathrm{T}} \boldsymbol{\delta}(\bold{s}_i, \bold{s}_j) - 1)\right]  \right).
\label{eq:log_loss_2}
\end{equation}
The gradient of the loss function $l_k$ is given by
\begin{equation}
\frac{\partial l_k}{\partial \bold{w}}=  \beta \sum_{(i,j) \in \mathcal{P}_k} h_{\beta} \left(  o_{ij} (\bold{w}^{\mathrm{T}} \boldsymbol{\delta}(\bold{s}_i, \bold{s}_j) - 1) \right) \boldsymbol{\delta}(\bold{s}_i, \bold{s}_j).
\label{eq:gradient_log_loss_2}
\end{equation}
Guided by the similarity of the proposed formulation with a logistic regression problem, we apply a classical inverse frequency weighting on the training pairs to compensate for class imbalance and we solve Eq.~\ref{eq:objective_function} with the L-BFGS algorithm, where a log transformation is used to parametrize the positive weights $\bold{w}$. The weights $\bold{w}$ in Eq.~\ref{eq:objective_function} are constrained not to be lower than $\bold{w}_1$ in the optimisation to ensure that the overlap probability does not increase in comparison to the standard bag of words model: ensuring this nonincreasing aspect was found to be essential in practice to avoid the repetition of negative suggestions. The optimisation strategy is guaranteed to converge to a global minimum due to the strict convexity of the objective function~\citep{mignon2012cvpr}.

\begin{figure*}[t]
\centering
\subfloat{\includegraphics[width=0.24\textwidth]{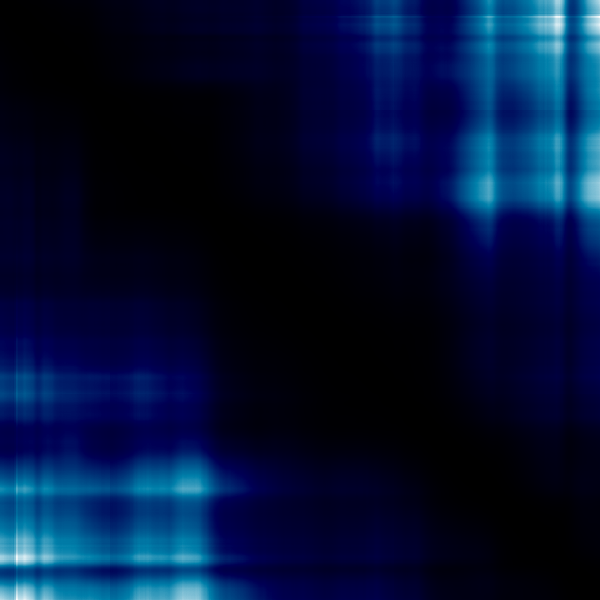}}\hfill
\subfloat{\includegraphics[width=0.24\textwidth]{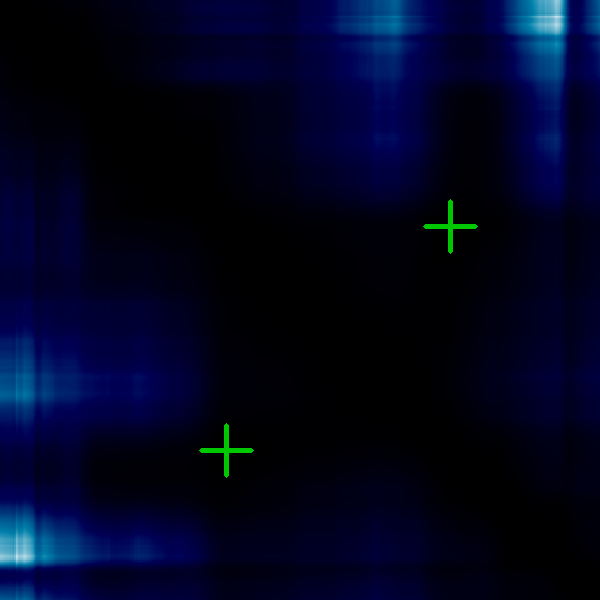}}\hfill
\subfloat{\includegraphics[width=0.24\textwidth]{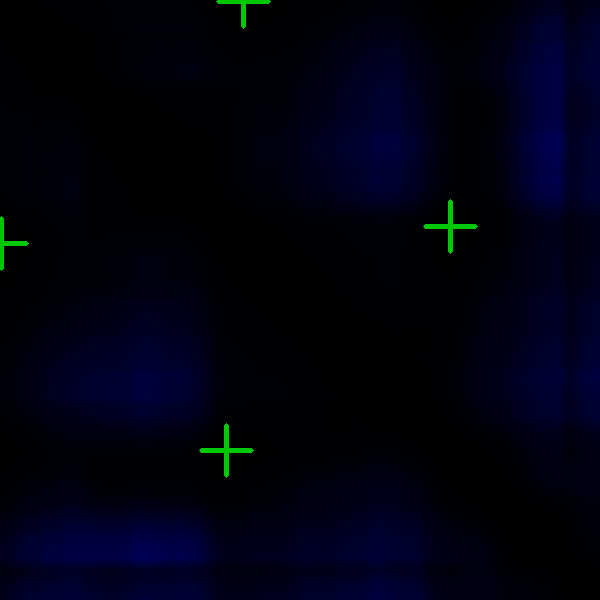}}\hfill
\subfloat{\includegraphics[width=0.24\textwidth]{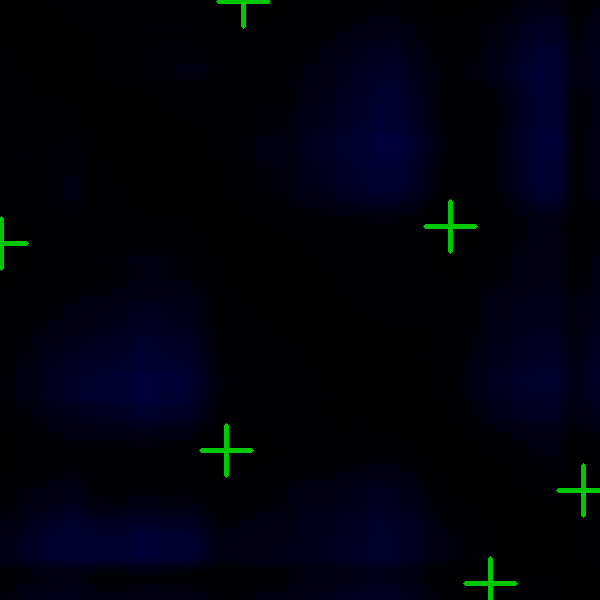}}\hfill
\subfloat{\includegraphics[width=0.24\textwidth]{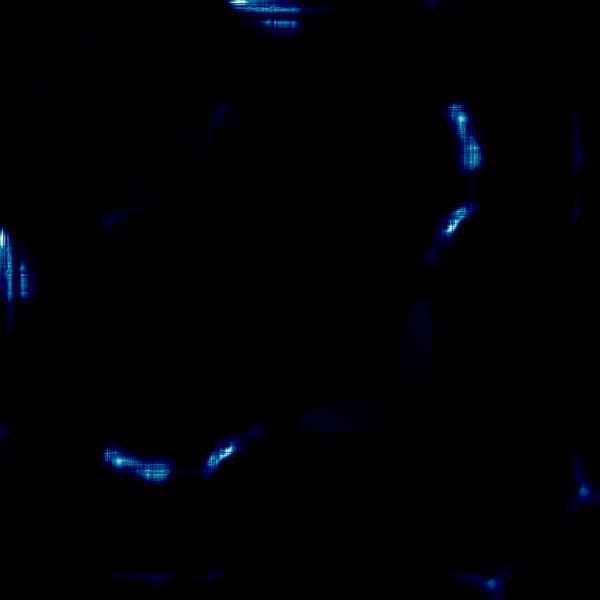}}\hfill
\subfloat{\includegraphics[width=0.24\textwidth]{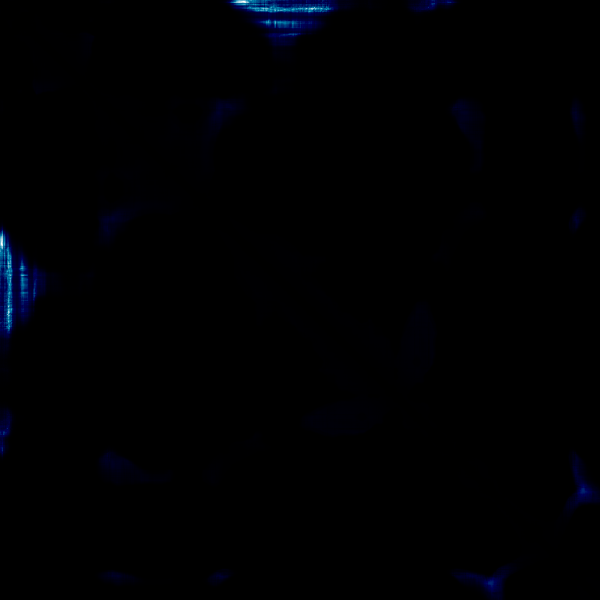}}\hfill
\subfloat{\includegraphics[width=0.24\textwidth]{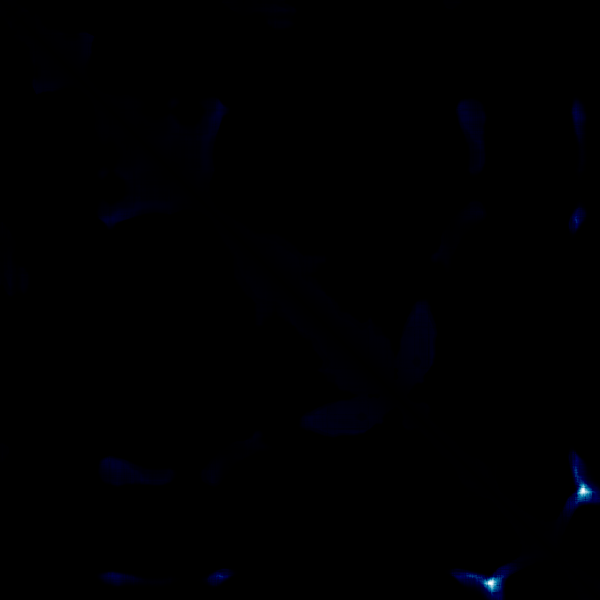}}\hfill
\subfloat{\includegraphics[width=0.24\textwidth]{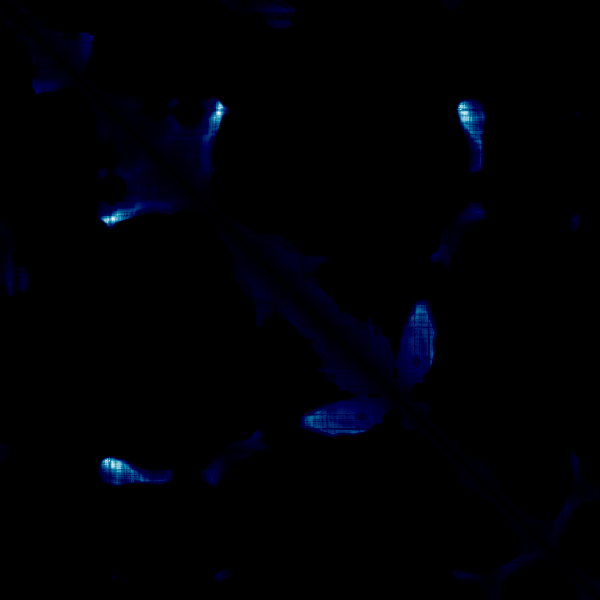}}\hfill
\caption{\textbf{Evolution of informativeness measure with added pairwise correspondences.} The first row displays the evolution of the informativeness measure as long-range overlapping pairs are annotated. A general decrease of the uncertainty is observed. The second row displays the expected reward at each iteration, i.e. where the probability of overlap is taken into account. For this second row, each map is individually normalised for visualisation.}
\label{fig:evolution_informativeness}
\end{figure*}

\subsection{Informativeness Measure}
\label{sec:annotation_reward}

In this section, we describe our choice of annotation reward $R_{ij}(\mathcal{T})$. The reward builds on an uncertainty measure $U_{ij}(\mathcal{T})$ stating the informativeness of collecting annotations on a pair of images $I_i$ and $I_j$ given the currently available correspondences and their uncertainty. Intuitively, the more uncertain the relative position between $I_i$ and $I_j$, the higher is the reward of having their pairwise correspondence revealed by the agent.
However, as we discussed in Section~\ref{sec:problem_statement}, a correspondence can only be revealed if $I_i$ and $I_j$ overlap, hence the consideration of the expected reward
\begin{equation}
\mathbb{E}[R_{ij}] =
 P^{\textrm{ext}}(O_{ij} \mid I_i, I_j, \bold{w}) P^{\textrm{pos}}(O_{ij} \mid \mathcal{T}) U_{ij}(\mathcal{T})
\label{eq:tradeoff_reward}
\end{equation}
instead, as defined in Eq.~\ref{eq:proba_overlap} and Eq.~\ref{eq:expected_reward}. We remind that $R_{ij}$ is defined as $U_{ij}$ if the two frames overlap and as $0$ otherwise.

A fundamental tradeoff is apparent on Eq.~\ref{eq:tradeoff_reward} as the uncertainty on the relative position between $I_i$ and $I_j$ increases. On the one hand, the informativeness $U_{ij}$ increases as we just discussed, but on the other hand, the position-based probability of overlap $P^{\textrm{pos}}(O_{ij} \mid \mathcal{T})$ decreases as a consequence of the increasing positional uncertainty. Asymptotically, this leads to an indeterminate form $0 \times \infty$ in Eq.~\ref{eq:tradeoff_reward}. This tradeoff is fundamental in the retrieval of long-range correspondences in mosaicking and was already identifed in the seminal work of \citet{Sawhney1998eccv} on topology inference. 

The importance of the choice of uncertainty measure is now apparent. Asymptotically, the behaviour of the system can be very different depending on whether the position-based overlap probability or the annotation reward dominates. Among the possible and intuitively appropriate uncertainty measures, we propose to set our uncertainty measure as to resolve the aforementioned indeterminate form such that $P^{\textrm{pos}}(O_{ij} \mid \mathcal{T}) U_{ij}(\mathcal{T})$ stays bounded, i.e. such that the two opposite effects asymptotically balance each other. As a consequence, the external overlap model $P^{\textrm{ext}}(O_{ij} \mid I_i, I_j, \bold{w})$ is asymptotically the dominant factor in Eq.~\ref{eq:tradeoff_reward}. This property is desired as it allows the use of external information regardless of the respective timepoints of $I_i$ and $I_j$ in the sequence, which is exactly the original purpose of this external overlap model.

To define our reward in such a manner, we study the asymptotic behaviour of $P^{\textrm{pos}}(O_{ij} \mid \mathcal{T})$ when the uncertainty increases. To do so, we exploit the bounds derived in Eq.~\ref{eq:bounds_position_probability} where the positional uncertainty is encoded in the eigenvalues $\nu_1$ and $\nu_2$ of the covariance matrix $\boldsymbol{\Sigma}_{\boldsymbol{\gamma}_{ij}}$ of the displacement between the centres of $I_i$ and $I_j$.
Since $\erf(t) = \frac{2}{\sqrt{\pi}} t + o(t) $ when $t \rightarrow 0$, we have for a fixed $x$ the asymptotic result
\begin{equation}
f_a(x,\nu) = \sqrt{\frac{2}{\pi \nu}}  a + o\left( \frac{1}{\sqrt{\nu}}\right)
\label{eq:asymptotic_result}
\end{equation}
when $\nu \rightarrow +\infty$, where $f_a$ was defined in Eq.~\ref{eq:function_f_overlap}.
Since $\hat{\bold{d}}_{ij} \cdot \vec{u}_1$ and $\hat{\bold{d}}_{ij} \cdot \vec{u}_2$ stay bounded by $\Vert \hat{\bold{d}}_{ij} \Vert$  when the uncertainty increases, we apply Eq.~\ref{eq:asymptotic_result} to both Eq.~\ref{eq:lower_bound_position_probability} and Eq.~\ref{eq:upper_bound_position_probability} and obtain, asymptotically, the approximate (due to the linearisation in Eq.~\ref{eq:linear_approximation_centre_distribution}) inequality
\begin{equation}
\frac{2 a_{\textrm{in}}}{\pi \sqrt{\nu_1 \nu_2}} \leq P^{\textrm{pos}}(O_{ij} \mid \mathcal{T}) \leq \frac{2 a_{\textrm{out}}}{\pi \sqrt{\nu_1 \nu_2}} 
\label{eq:asymptotic_bounds}
\end{equation}
when the eigenvalues $\nu_1, \nu_2 \rightarrow +\infty$. Thus the position-based overlap probability decreases like $\frac{1}{\sqrt{\nu_1 \nu_2}}$. Defining the informativeness measure as $U_{ij}(\mathcal{T}) = \sqrt{\nu_1 \nu_2}$ exactly compensates this effect by leading to the asymptotic bounds
\begin{equation}
\frac{2 a_{\textrm{in}}}{\pi} \leq P^{\textrm{pos}}(O_{ij} \mid \mathcal{T}) U_{ij}(\mathcal{T}) \leq \frac{2 a_{\textrm{out}}}{\pi}
\label{eq:asymptotic_bounds_practical_reward}
\end{equation}
when $\nu_1, \nu_2 \rightarrow +\infty$. Since the product of the eigenvalues of a matrix is its determinant, we have just defined
\begin{equation}
U_{ij}(\mathcal{T}) = \sqrt{\det  \boldsymbol{\Sigma}_{\boldsymbol{\gamma}_{ij}}},
\label{eq:def_reward}
\end{equation}
i.e. we take as uncertainty measure the generalised standard deviation of the multivariate Gaussian distribution governing $\boldsymbol{\gamma}_{ij}$.

\section{Baselines}
\label{sec:baselines}

Before describing our experiments and results, we summarise in this section the baselines based on previous works that we implemented for our experiments and highlight the differences with our method.

\subsection{\citet{Elibol2013ras}}
\label{sec:elibol}

The topology inference method by \citet{Elibol2013ras} is the work which contains the most similarities with our approach, as it also introduces a probabilistic overlap model based on covariance propagation (but intrinsically aimed at short-range correspondences only, as discussed below), and an uncertainty measure to prioritise the registration of pairs.

The probabilistic overlap model proposed in \citet{Elibol2013ras} is introduced in the case of transformations with only $4$ degrees of freedom and is based on a different bundle adjustment formulation. The key difference with our framework resides in the fact that their probabilistic model is first used as a filtering part, where only the pairs with over $99\%$ of probability to overlap are retained as candidates. As a second step, for each candidate, a Monte Carlo estimation is run to refine further the set of candidates. Adapting the idea to our context, we implement this baseline by filtering out, at each iteration, the pairs for which the upper bound on the overlap probability (as defined in Eq.~\ref{eq:bounds_position_probability}) is lower than $0.99$. For the retained pairs, the overlap probability is computed with Monte Carlo sampling the same way as in our model.

In \citet{Elibol2013ras}, the informativeness of each candidate pair is defined as the reduction of the entropy of the joint covariance matrix of the bundle adjustment results (in our case, defined in Eq.~\ref{eq:covariance_propagation_haralick}). Unfortunately, this computation is too costly in an interactive setting, as it requires an update of the bundle adjustment for each candidate pair. After communication with the authors, we followed their recommendation of approximating this entropy as done in one of their earlier works~\citep{elibol2010augmented}, which leads to the approximate uncertainty measure
\begin{equation}
U_{ij}^{\textrm{Elibol}}(\mathcal{T}) = \log \det \boldsymbol{\Sigma}_{\boldsymbol{\gamma}_{ij}}.
\label{eq:def_reward_elibol}
\end{equation}
This approximation corresponds to computing the entropy of the displacement $\boldsymbol{\gamma}_{ij}$ which is tractable. 

We can note that, unlike our informativeness measure defined in Section~\ref{sec:annotation_reward}, the uncertainty measure of Eq.~\ref{eq:def_reward_elibol} yields a vanishing expected reward as the uncertainty increases. Through their choice of informativenes measure and their threshold on the overlap probability to reduce false positives, the approach of \citet{Elibol2013ras} is thus, by design, not aimed at the retrieval of long-range correspondences. Instead, it follows a more conservative strategy where short-range correspondences are progressively found and included in the bundle adjustment. Although this strategy has the advantage of reducing the amounts of false positive suggestions, a larger annotation budget is required to reach convergence of the reconstruction, as will be shown in our experiments. In addition, it relies on the hypothesis that short-range correspondences are close enough to be retrieved, and is thus unable to handle sequences where overlapping frames are at too distant timepoints.

\subsection{\citet{Sawhney1998eccv}}

The seminal work of \citet{Sawhney1998eccv} first identified the fundamental tradeoff between probability of overlap and informativeness of image pairs. In their work, the probability of frame overlap is entirely defined on the reconstructed mosaic at a given iteration, without considering its uncertainty. The probability of overlap is encoded by an arc length defined as
\begin{equation}
l_{ij} = \frac{\max(0,\Vert \hat{\boldsymbol{\gamma}}_{ir} - \hat{\boldsymbol{\gamma}}_{jr} \Vert - |R_i - R_j| )}{\min(2 R_i, 2 R_j)},
\label{eq:sawhney_overlap_score}
\end{equation}
where $R_i$ is the radius of the image $I_i$ warped on the reference frame. More precisely, this radius is defined as the mean distance of the warped corners to the warped centre $\hat{\boldsymbol{\gamma}}_{ir}$ of the image. If $l_{ij} \geq 1$, the two frames do not overlap. The lower $l_{ij}$ is, the likelier are the two frames to overlap.

To define their informativeness measure, \citet{Sawhney1998eccv} consider at each iteration $k$ a weighted graph of overlapping pairs $\mathcal{G}_k$ defined as follows. Each vertex of $\mathcal{G}_k$ is a frame, and two frames are connected by an edge if and only if they have been previously identified as overlapping at the previous iterations. The weight of each edge connecting a frame $I_i$ and a frame $I_j$ is defined as the arc length $l_{ij}$ defined above. The informativeness of registering a pair $(I_i,I_j)$ is then defined as the length $L_{ij}$ of the shortest path between $I_i$ and $I_j$ in the graph $\mathcal{G}_k$.
 
In their paper, \citet{Sawhney1998eccv} suggest to attempt the registration of a pair $(I_i,I_j)$ if $l_{ij}$ when ``\textit{$l_{ij}$ does not exceed a certain limit and is significantly lower than $L_{ij}$}''. We implement this idea by considering the probability
\begin{equation}
P^{\textrm{Sawhney}}(O_{ij} \mid \mathcal{T}) = \max(0,1-l_{ij}),
\label{eq:sawhney_probability}
\end{equation} 
which encourages the suggestions of pairs of low values of $l_{ij}$, and the informativeness measure
\begin{equation}
U_{ij}^{\textrm{Sawhney}}(\mathcal{T}) = \max \left( 0, \frac{L_{ij}} {l_{ij}} - 1\right)
\end{equation}
which rewards large relative differences between $l_{ij}$ and $L_{ij}$.
Considering the ratio between $L_{ij}$ and $l_{ij}$ is also what \citet{Marzotto2004cvpr} proposed to implement the approach presented by \citet{Sawhney1998eccv}. Note that, after registering a pair $(I_i,I_j)$, we have $L_{ij} \leq l_{ij}$ by definition of $L_{ij}$ and thus the resulting informativeness of suggesting $(I_i,I_j)$ again is then $0$, as desired.

%

\begin{figure*}[t]
\centering
\subfloat[True trajectory]{\includegraphics[width=0.95\textwidth]{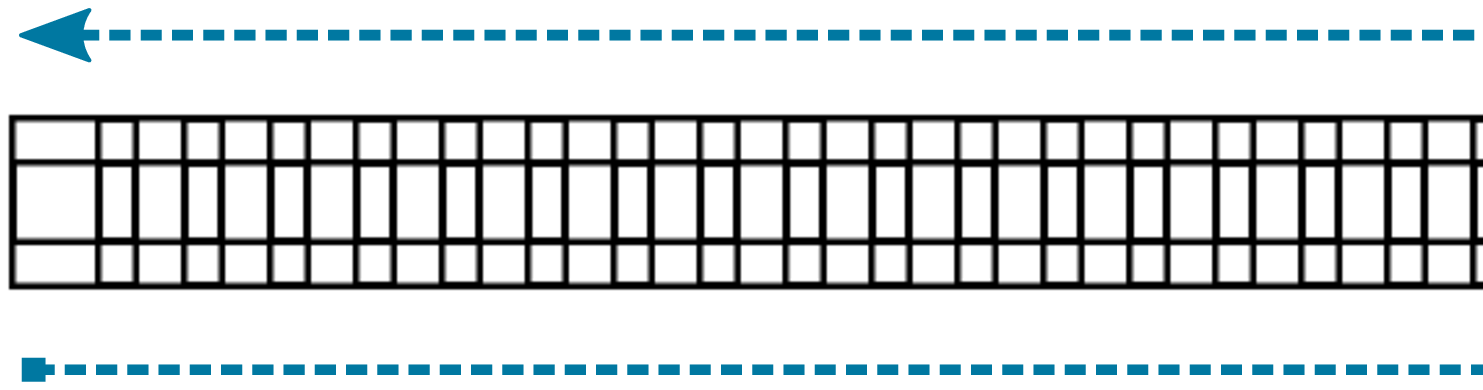} \label{fig:true_raster_trajectory}}\hfill
\subfloat[Initial reconstruction]{\includegraphics[width=0.95\textwidth]{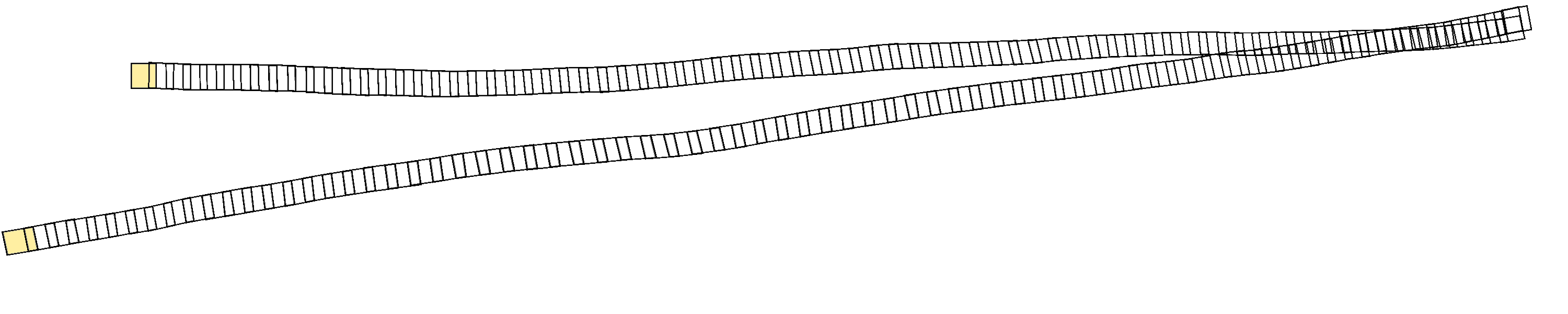} \label{fig:drift_raster_trajectory}}\hfill
\caption{\textbf{Synthetic raster scan.} (a) Example of the considered raster scan trajectory, where the camera moves towards the right and then back to the left after a slight vertical shift upwards. All frames sharing the same horizontal coordinate overlap. (b) Actual observed reconstruction when registering consecutive frames. Due to drift, the most informative pair of frames (marked in yellow) cannot be identified as overlapping if one only relies on this initial reconstruction.}
\label{fig:raster_trajectories}
\end{figure*}

\section{Experiments}
\label{sec:experiments}

In this section, we present an experimental validation of our method. We compare our approach with the two baselines described in Section~\ref{sec:baselines}. In addition, we also consider restricted variants of our approach where only one of the two complementary overlap models is used instead of the combination of both. These additional baselines allow us to assess the individual contribution of each overlap model. 

To quantitatively evaluate the performance of the implemented methods, we rely on preliminarily acquired gold standard correspondences on some pairs of frames in the considered sequence. The procedure to acquire these correspondences depends on the dataset and is described in more details below where each individual dataset is presented. Given ground truth correspondences on a sequence, the performance of a method at a given iteration is assessed by reporting the mean relative root-mean-square-deviation (RMSD) of pairwise landmark errors, i.e. we measure how the reconstructed mosaic matches the available ground truth landmarks. Note that we compute the pairwise reconstruction errors in the reference frame of each pair, such that our evaluation measure does not depend on the chosen reference frame in which the mosaic is reconstructed.

We implemented\footnote{Code available at https://github.com/LoicPeter/video-mosaicking} our algorithm in C++ using the OpenCV library~\citep{opencv_library}.
All experiments were run on an Intel\textsuperscript{\textregistered} Core\textsuperscript{TM} i7-8650U @ 1.90GHz laptop with 8 CPUs. In this setup, the waiting time between two interactions does not exceed a few seconds.

Our evaluation is conducted on three datasets. First, we consider synthetic trajectories to experimentally validate on simple cases the behaviours and theoretical properties of the various methods (Section~\ref{sec:experiments_synthetic}). Then, we consider a publicly available dataset on aerial imaging with a fully-automated agent, i.e. a scenario of bundle adjustment with automated pairwise registration (Section~\ref{sec:experiments_aerial}). Finally, we study an interactive scenario on an in vivo biomedical sequence with challenging visual conditions, where pairwise annotations are provided by a human user  (Section~\ref{sec:experiments_fetoscopy}).

\subsection{Synthetic Examples}
\label{sec:experiments_synthetic}

We first illustrate and verify the theoretical properties of our framework by considering two synthetic trajectories under simple settings. By construction, the true position of each frame on the mosaic is known and can thus be used for validation. For simplicity, in both cases, the true underlying transformation relating each pair of consecutive frames is restricted to a translation.

\subsubsection{Raster Scan}
\label{sec:raster_scan}

\paragraph{Experimental Design}

First, we consider a raster scan of $N = 1000$ frames, where the position of the centre of each frame $I_n$ is defined as
\begin{equation}
(x(n),y(n)) = ( n \delta_x, 0)
\end{equation}
if $n \leq \frac{N}{2}$, and 
\begin{equation}
(x(n),y(n)) = ((N - n + 1) \delta_x, \delta_y)
\end{equation}
otherwise. This corresponds to a camera moving along the $x$-axis in a straight line, then going up from $\delta_y$ between the timepoints $\frac{N}{2}$ and $\frac{N}{2} + 1$, and then moving back until it reaches its initial $x$-coordinate (Fig.~\ref{fig:true_raster_trajectory}). $\delta_x$ and $\delta_y$ are set to a third of the image dimensions $100 \times 100$. The measured pairwise registrations between overlapping frames are obtained by randomly adding a Gaussian error of mean $0$ and standard deviation $1$ on control points that are located within the overlapping area of the two frames. For the purpose of this first experiment, we consider an ideal external model which is perfectly representative of the amount of overlap between two frames: the external overlap probability between two images is set to the area of intersection of the images divided by the area of the image domain. We repeat this experimental setup $5$ independent times, where each run corresponds to different randomly drawn errors on the pairwise registrations.

This raster scan trajectory is such that any pair of the form $(I_n,I_{N-n})$ overlaps. Moreover, the lower $n$, the more informative such a pair is expected to be (before any interaction with the agent). However, due to the large amount of frames, these informative long-range pairs cannot be retrieved by only considering the initial, drift-affected reconstructed mosaic (Fig.~\ref{fig:drift_raster_trajectory}). Therefore, by design, this synthetic example illustrates the advantage of the asymptotic guarantees made possible by the choice of our overlap probability and uncertainty measure.

\paragraph{Results}

\begin{figure}[t]
\centering
\includegraphics[width=0.40\textwidth]{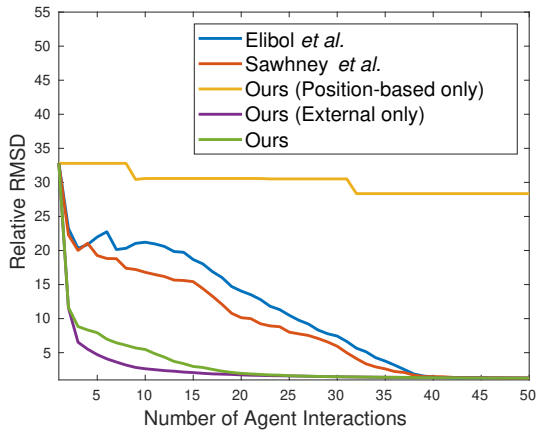}
\caption{\textbf{Results on a synthetic raster scan trajectory.} Under an ideal external model of frame overlap, the asymptotical properties of our approach allow a quick retrieval of the most informative pairs (hence a faster convergence), although these highly informative pairs are not considered as overlapping do not overlap in the initial reconstruction due to drift. As prior works by \citet{Sawhney1998eccv} and \citet{Elibol2013ras} rely on this initial reconstruction to predict the overlap between frames, these methods are only able to retrieve pairs at a shorter range and thus converge slower.}
\label{fig:results_raster}
\end{figure}

We report the obtained RMSD for up to $50$ interactions in Fig.~\ref{fig:results_raster}. The reported RMSD is averaged over the $5$ runs yielding different pairwise registrations. We observe that the strategies of \citet{Sawhney1998eccv} and \citet{Elibol2013ras} both cannot retrieve the most informative pairs at once: indeed, due to the amount of drift, the overlap score of these most informative pairs is \textit{exactly} $0$ and these pairs can, therefore, not be retrieved. Instead, we observe a steady decrease in RMSD which corresponds to ``mid-range'' overlapping pairs being progressively retrieved, yielding a progressive recovery of the true trajectory. 

In contrast to these baselines, our approach is able, thanks to the asymptotical balance between position-based overlap model and uncertainty measure (Section~\ref{sec:annotation_reward}), to exploit the information contained in the external overlap model to directly retrieve the most informative pairs. As expected, the importance of an external model in our approach is established by the low performance of the ``Position-based only'' strategy. Note that, in this first theoretical case, the assumption of an ideal external model no longer justifies the need for a position-based overlap model at all: the ``External only'' baseline has access to all the necessary information about the trajectory and rather acts as an oracle baseline.

%
%
%

\subsubsection{Circular Trajectory}

\paragraph{Experimental Design}

\begin{figure}[t]
\centering
\subfloat[True trajectory]{\includegraphics[width=0.20\textwidth]{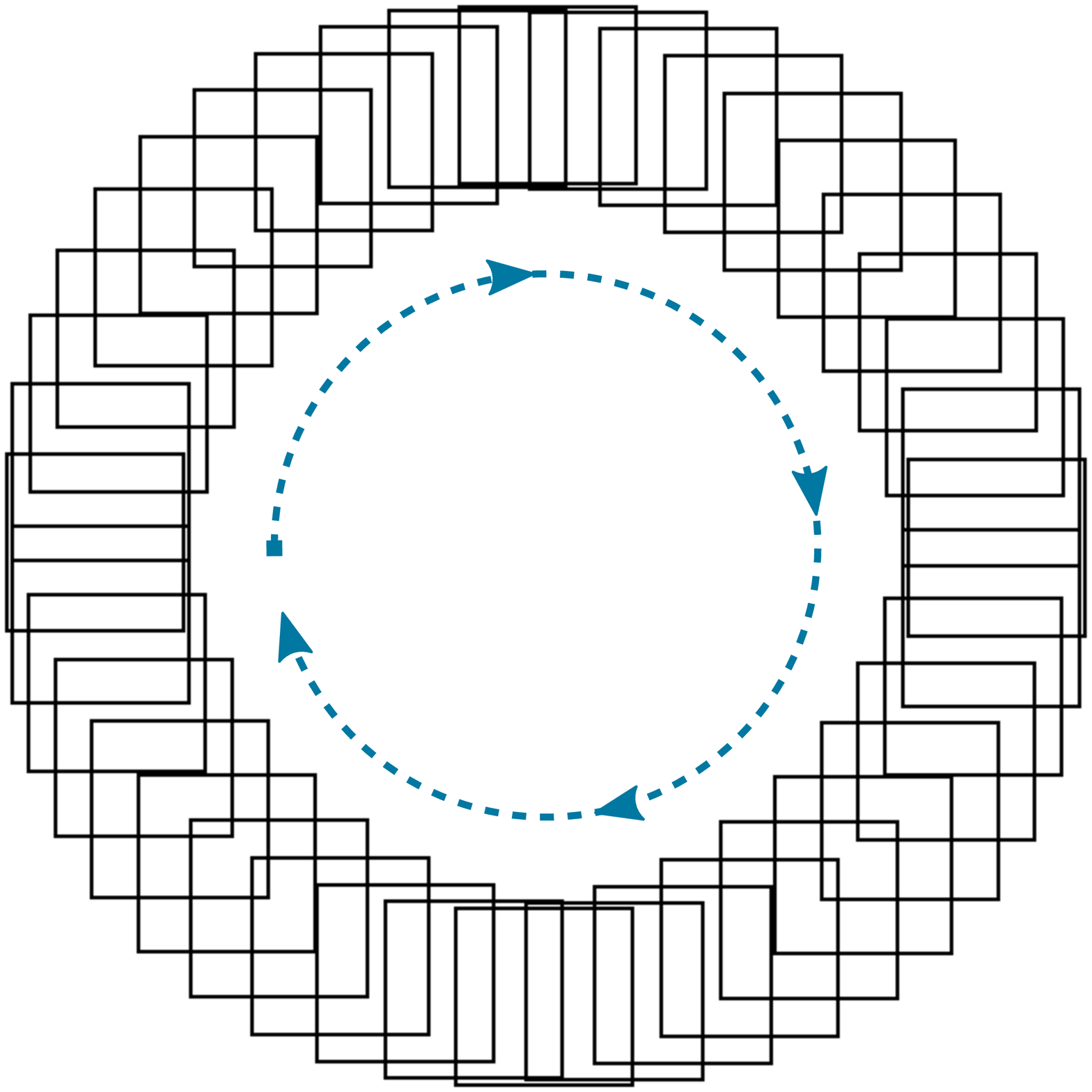}}\hfill
\subfloat[Initial reconstruction]{\includegraphics[width=0.273\textwidth]{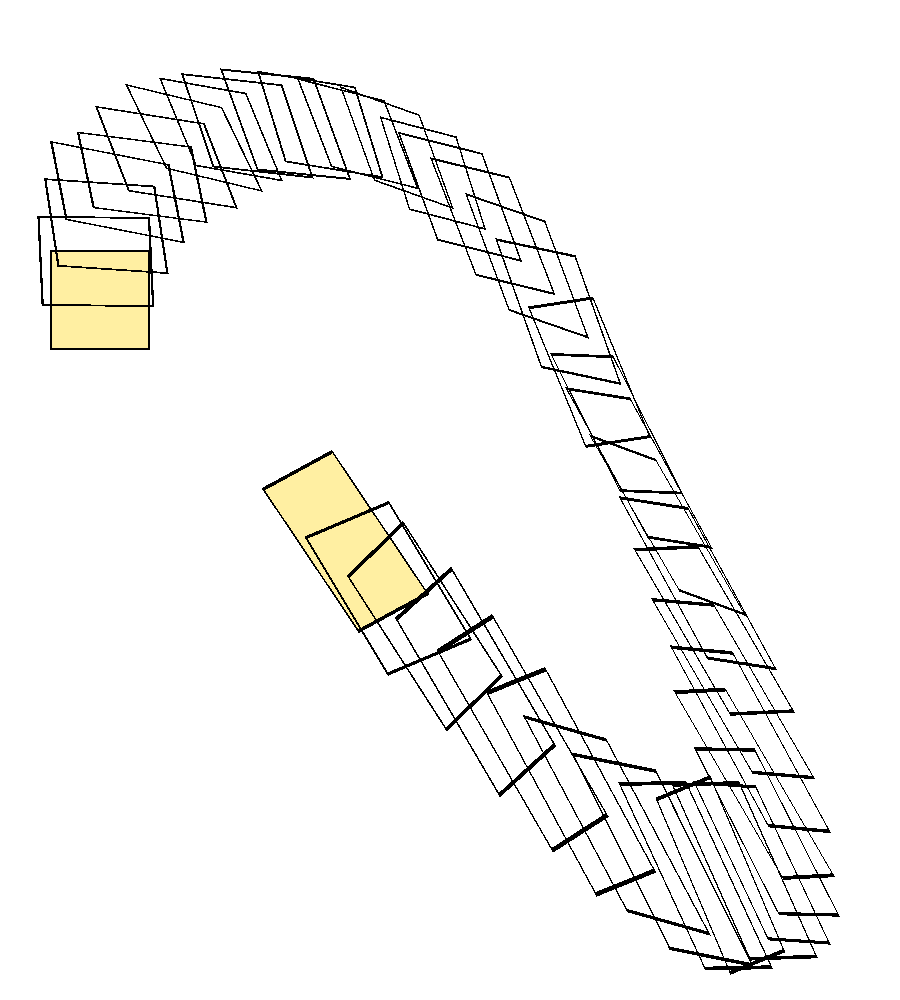}}\hfill
\caption{\textbf{Synthetic circular trajectory.} (a) Considered trajectory consisting of one revolution. (b) Initial reconstruction obtained when registering consecutive frames.}
\label{fig:circular_trajectories}
\end{figure}

We consider another synthetic example consisting of a circular trajectory with one revolution: the coordinates of the true position of each frame $I_n$ are defined as
\begin{equation}
(x(n),y(n)) = \left( R \cos (\frac{2 \pi n}{N}), R \sin (\frac{2 \pi n}{N})\right),
\end{equation}
with $N = 1000$ and $R = 250$. Due to the amount of drift, the overlapping pairs which are able to close the loop and reconstruct a drift-free mosaic, such as $(I_1, I_{1000})$, are not immediately retrievable if one only considers the initial reconstructed mosaic (Fig.~\ref{fig:circular_trajectories}). Similarly to the raster scan scenario, we consider $5$ runs with varying random pairwise registrations as measurements.

For this synthetic case, we consider a more challenging external overlap model as the ideal model designed in the raster scan scenario. We consider that the signature of each frame is given by
\begin{equation}
\textbf{s}^{(n)} = \left(\cos (\frac{4 \pi n}{N}) , \sin (\frac{4 \pi n}{N}) \right),
\end{equation}
i.e. encodes the position of the frame by its angle modulo $\pi$. As a result, diametrically opposite frames are wrongly considered as likely to overlap by this external model. 

\paragraph{Results}

\begin{figure}[t]
\centering
\includegraphics[width=0.40\textwidth]{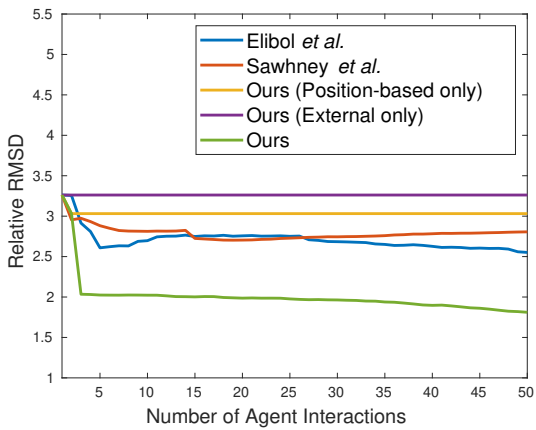}
\caption{\textbf{Results on a synthetic circular trajectory.} On the constructed circular trajectory, the pairs allowing to close the loop cannot be found based on the initial reconstruction due to drift. As no ``medium-range'' overlapping pairs are present in the circular trajectory, the baselines are unable to retrieve any informative overlapping pair. Under the assumption of a partially erroneous external model, the results also emphasise the importance of incorporating a complementary position-based model of frame overlap.}
\label{fig:results_circular_wrong}
\end{figure}

The results in this setting (Fig.~\ref{fig:results_circular_wrong}) show that our approach quickly retrieves the long-range overlapping pair. The results also illustrate the value of using the position information as soon as the external model is error-prone. Here, for diametrically opposite frames, the position-based overlap model is ``sufficiently certain'' about the non-overlap of these frames (i.e., the asymptotic regime is not reached yet) to mitigate the erroneous external model which would otherwise encourage to retrieve these non-overlapping pairs.

Unlike the raster scan scenario considered above, the topology of the circular trajectory results in the impossibility for previous approaches to retrieve the long-range pairs. Indeed, in this case, the absence of ``medium-range'' overlapping pairs does not enable the progressive recovery of the topology of the sequence.

%

\subsection{Aerial Imaging Sequence}
\label{sec:experiments_aerial}

\paragraph{Experimental Setup} 

We evaluate our approach on the publicly available sequence introduced by \citet{Xia2017pr}, which consists of $744$ aerial views of an urban area, acquired along a raster scan of $24$ strips. The good contrast conditions in the sequence enable the reliable use of a classical automated keypoint-based registration combining SURF and RANSAC. We accordingly consider an automated agent and evaluate our approach in a bundle adjustment context, where the objective is the reduction of the number of matching attempts, as in the considered previous works~\citep{Elibol2013ras,Sawhney1998eccv}. Our external overlap model is built as described in Section~\ref{sec:content_based_overlap_probability} by computing the bag of words descriptors on salient locations extracted with SURF within each frame, without registration. 

\paragraph{Results}

\begin{figure}[t]
\centering
\includegraphics[width=0.4\textwidth]{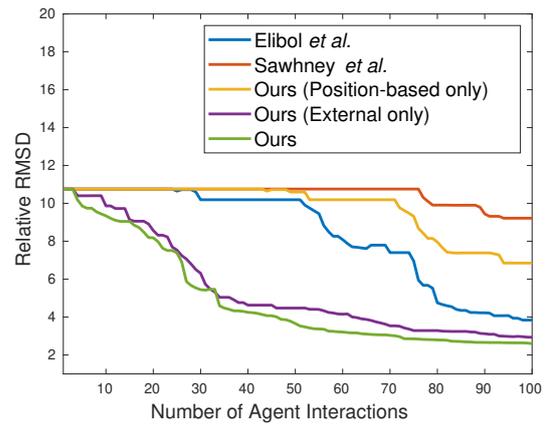}
\caption{\textbf{Results on an aerial imaging sequence.} On this real-world sequence, the agent is an automated pairwise matching algorithm. By selecting the pairs to be matched according to their expected reward, our approach yields the fastest reduction of the bundle adjustment reconstruction error. The number of attempted pairwise registrations to reach a certain bundle adjustment accuracy is thus reduced in comparison to the two baselines which were introduced to address the same problem.}
\label{fig:results_aerial}
\end{figure}

\begin{figure*}[t]
\centering
\includegraphics[width=0.95\textwidth]{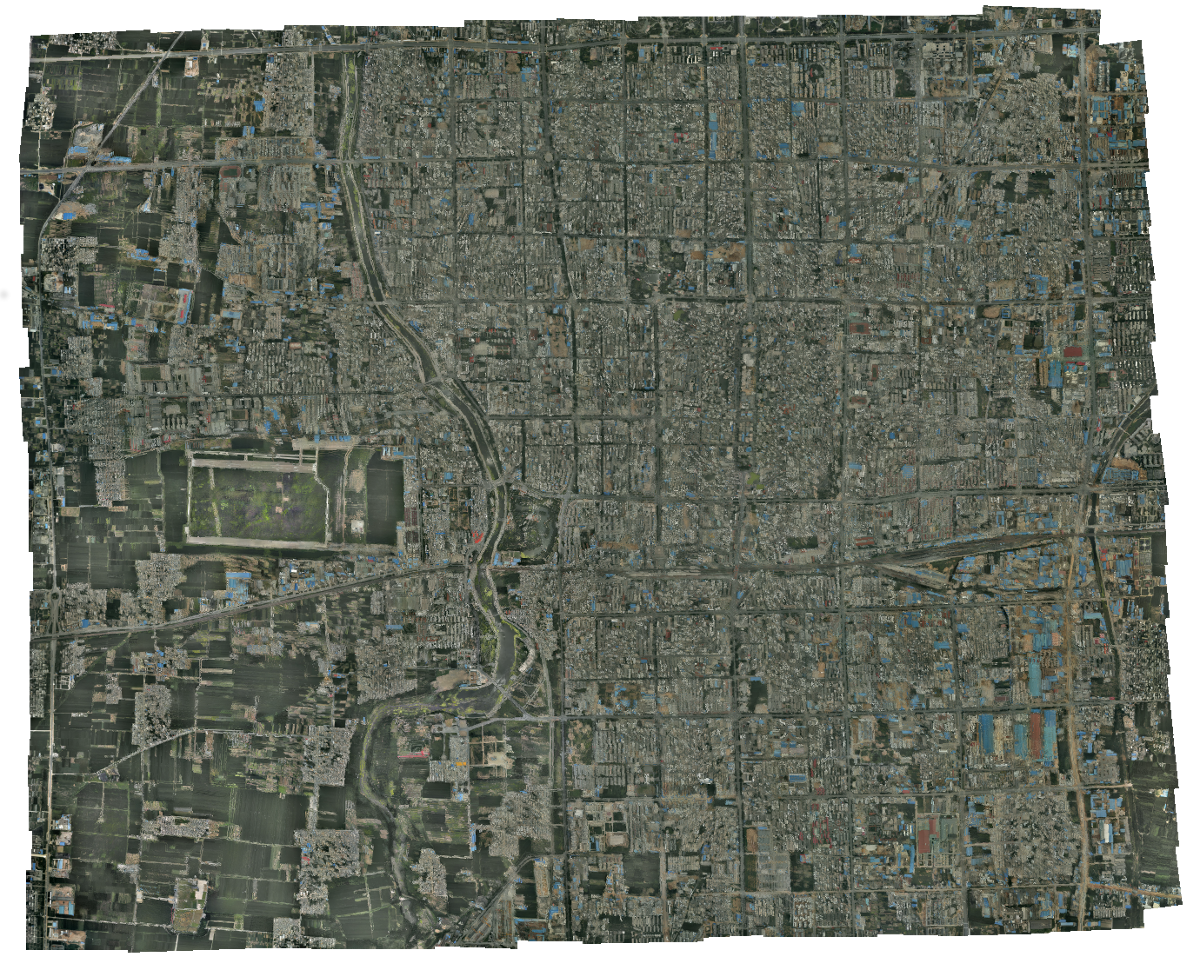}
\caption{\textbf{Reconstructed aerial mosaic after $100$ automated long-range queries}. After attempting the pairwise registration of what only represents $3.24\%$ of the overlapping pairs, our approach yields a globally consistent mosaic. For comparison, the initial drift-affected reconstruction obtained by only registering consecutive frames can be seen in Fig.~\ref{fig:aerial_initial_mosaic}.}
\label{fig:aerial_final_mosaic}
\end{figure*}

We report quantitative results in Fig.~\ref{fig:results_aerial}.
Similarly to our synthetic raster scan, overlapping pairs of frames are sufficiently close in the sequence to be retrieved based on a reconstructed mosaic. The approach of \citet{Elibol2013ras} which prioritise conservative matches and follows an entropy-based scheduling converges faster than the model of \citet{Sawhney1998eccv}, for which the less conservative nature of the overlap model results in a higher number of false positives and a slower convergence. Our approach demonstrates overall a faster convergence than these two baselines. We show in Fig.~\ref{fig:aerial_final_mosaic} the reconstructed mosaic obtained with our approach after $100$ iterations, which represents only $3.26\%$ of the total number of overlapping pairs (excluding the registered consecutive frames). In spite of the relatively small number of added correspondences, the mosaic displays a strong spatial consistency, and clearly improves over the drift-affected initial reconstruction shown in Fig.~\ref{fig:aerial_initial_mosaic}.

\subsection{Fetoscopy Sequence}
\label{sec:experiments_fetoscopy}

\paragraph{Experimental Setup}

Finally, we evaluate our framework on a medical sequence acquired to addresss a condition called Twin-to-Twin Transfusion Syndrome~\citep{baschat2011twin}. This condition occurs during pregnancy, when a twin set shares the same placenta with a given set of direct twin-to-twin vascular connections known as anastomoses which creates a blood imbalance between the twins. Since this condition is lethal, surgery is traditionally conducted by photocoagulating the anastomoses with a laser under direct endoscopic vision, the latter referred to as fetoscopy. However, the limited field of view of the endoscope results in an impractical navigation for the surgeon who requires external assistance via ultrasound guidance by a second operator. This causes long operation times and a lack of anatomical insight, hence inappropriate or incomplete selection of target vessels, in turn leading to incomplete surgery or overkill. It is therefore of high interest to the surgeon to be able to reconstruct a map of the placenta from an exploratory sequence~\citep{Reeff}, which could then be used for localisation and orientation during surgery.

We consider a sequence of $600$ frames acquired in in vivo conditions at the University College Hospital in London, UK. This sequence differs from the aerial dataset for two main reasons. First, the trajectory of the fetoscope is implicitly driven by the vascular network followed by the surgeon. This trajectory is not as structured as a raster scan, and is not known prior to acquisition. Second, the low contrast within the images (see Fig.~\ref{fig:example_low_range} and Fig.~\ref{fig:correspondences}) results in more challenging visual conditions and in the inapplicability of classical feature-based registration. This fact was previously illustrated in~\citet{Peter2018ijcars}, where we developed a dense registration technique suitable for this task and achieving reliable registration of consecutive frames. However, the challenging visual differences between long-range overlapping frames (as illustrated in Fig.~\ref{fig:example_low_range}) do not allow a systematic, reliable matching of long-range frames as the registration algorithm remains prone to failures on these pairs. As a result, a large number of matching attempts combined with conservative heuristics had to be conducted in our previous work~\citep{Peter2018ijcars}.

We study how our approach can facilitate, in an interactive manner, the reconstruction of the mosaic in such challenging visual conditions. In this scenario, the agent is thus a human annotator on whom we rely to annotate the suggested pairs of frames. To establish a ground truth on this dataset, since automated registration between long-range pairs is not reliable, we manually annotated a total of $133$ gold standard landmark correspondences on $60$ pairs of frames distributed over the sequence. The bag of words descriptors used in our external overlap model are here densely computed on a grid. 

\paragraph{Results}

Results are reported in Fig.~\ref{fig:results_fetoscopy}. We observe that the method of \citet{Sawhney1998eccv} is not able to retrieve overlapping frames and, on closer inspection, repeatedly suggests false negatives within the same area. This illustrates the importance of an updatable model which can be adjusted from false positives and avoid the repetition of similar erroneous suggestions. Our method and all other baselines yield a similar performance on this sequence. The strategy of \citet{Elibol2013ras} is successful here due to the fact that long-range overlapping frames are located at sufficiently close timepoints to not require asymptotic guarantees on the expected reward.

In terms of clinical application, the results are a clear improvement over our previous work~\citep{Peter2018ijcars}. 
In less than $10$ interactions (out of $179,700$ image pairs), we are able to reconstruct a mosaic of comparable qualitative accuracy as the one presented in~\citet{Peter2018ijcars}, for which all pairs above a similarity threshold of $0.8$ were inspected, leading to a total to $2025$ registration attempts. By retrieving a few well-chosen pairs instead, a $200$-fold improvement is thus achieved, and a sufficiently small number of required interactions is reached to be tractable for a human annotator. As a result, our approach also addresses the difficulty of quantitatively measuring the performance of mosaicking algorithms on sequences where feature-based registration is not feasible. Through the interactions with the annotator, reliable landmarks are collected on image pairs that are by nature informative and visually diverse, and can serve as gold standard annotations for future research on this sequence.

\begin{figure*}[t]
\centering
\subfloat[Sequential mosaicking]{\includegraphics[width=0.45\textwidth]{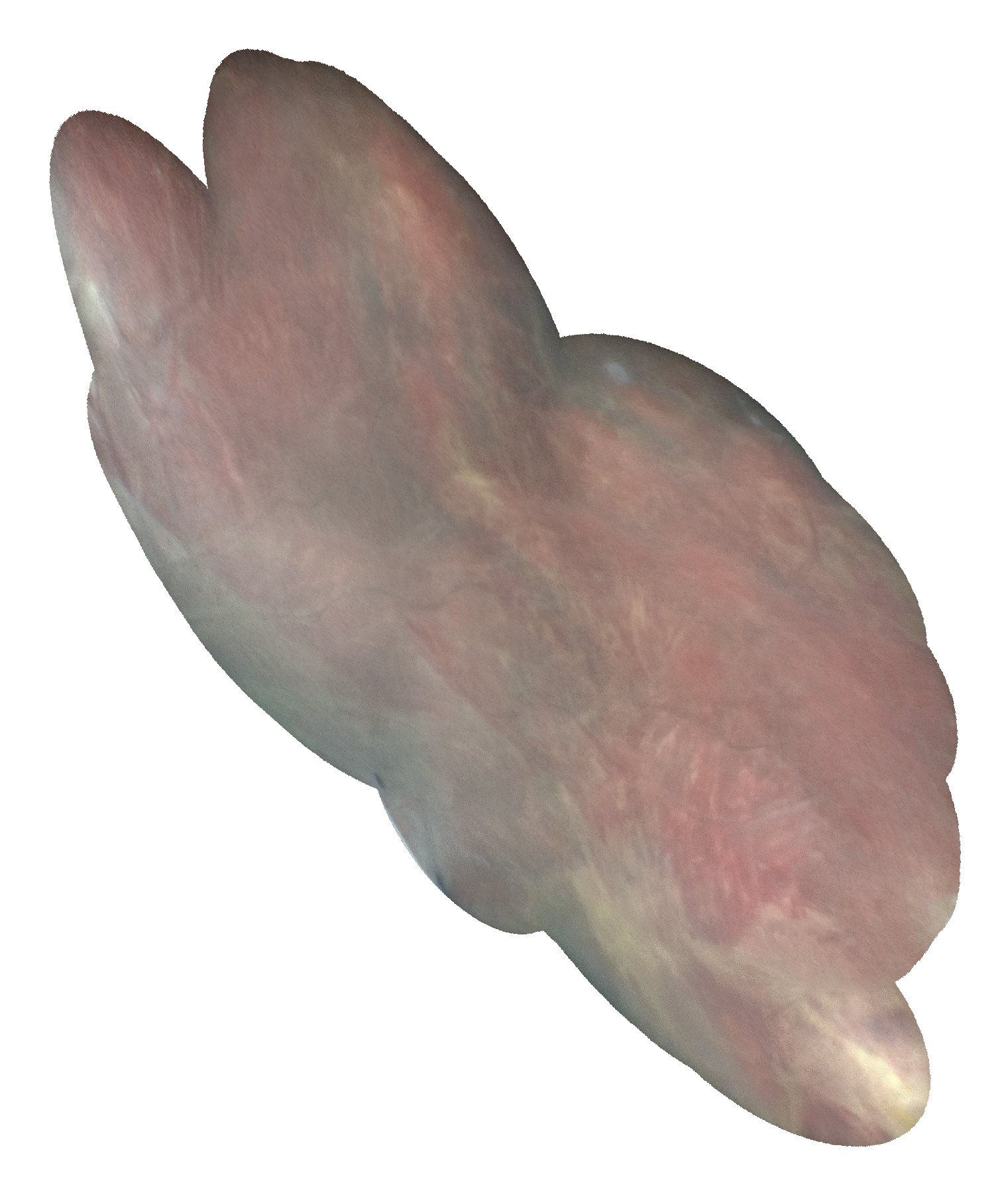}}\hfill
\subfloat[Reconstruction after $10$ interactions]{\includegraphics[width=0.55\textwidth]{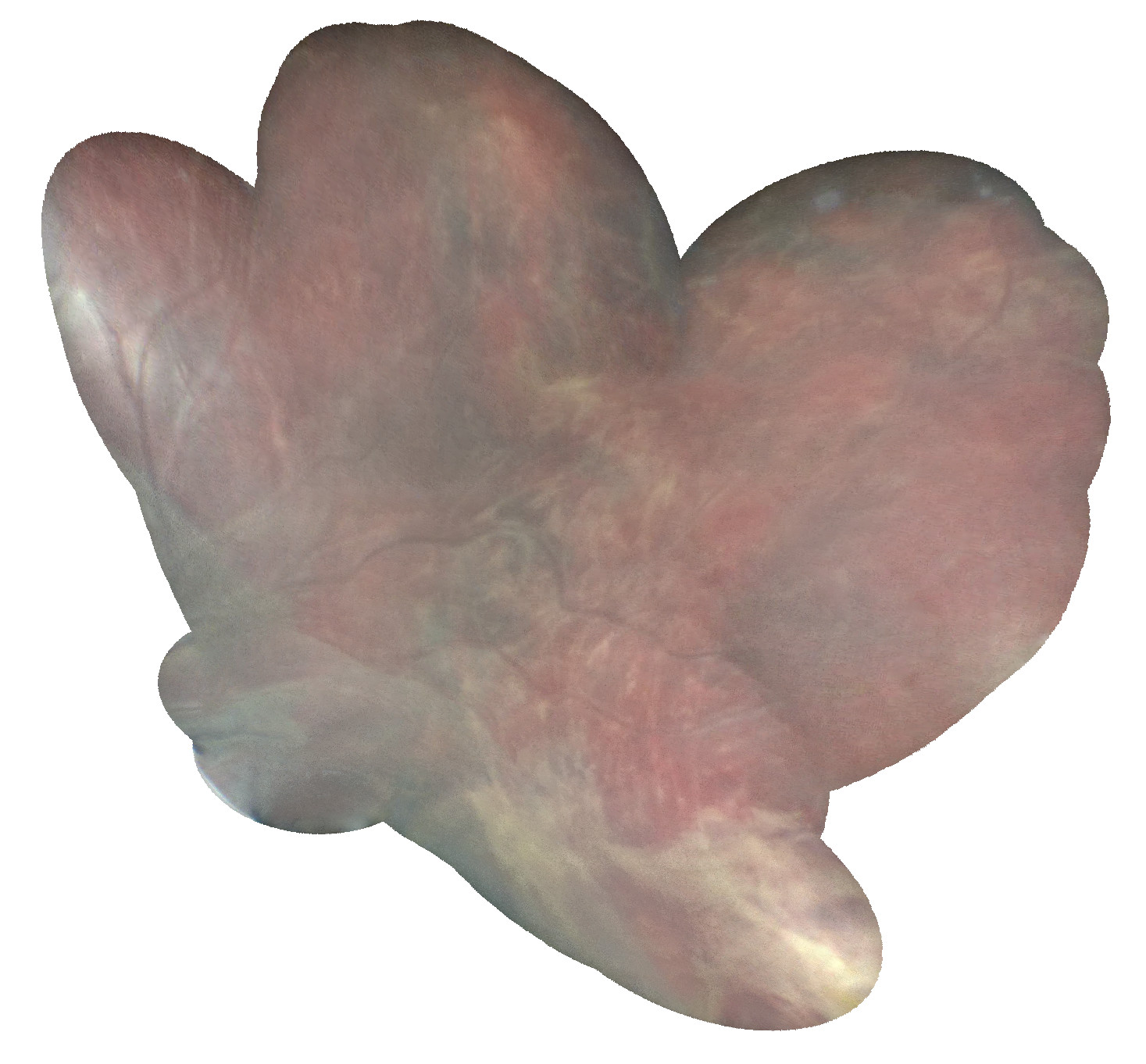}}\hfill
\caption{\textbf{Mosaics reconstructed from the fetoscopy sequence.} (a) Mosaic reconstructed based on sequential alignment. Drift can be observed. (b) After only $10$ interactions, a globally consistent mosaic can be reconstructed.}
\label{fig:mosaics_fetoscopy}
\end{figure*}

\begin{figure}[t]
\centering
\includegraphics[width=0.4\textwidth]{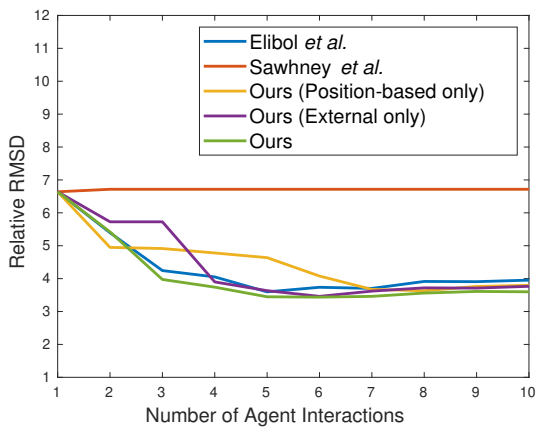}
\caption{\textbf{Results on a fetoscopy sequence.} In difficult low-contrast visual conditions where automated registration of long-range pairs is not fully reliable, a few well-chosen interactions with a human user can successfully reduce the reconstruction error of the initial drift-affected mosaic (see also Fig.~\ref{fig:mosaics_fetoscopy} for qualitative results).}
\label{fig:results_fetoscopy}
\end{figure}

\section{Conclusion}

We introduced an approach for retrieving informative long-range correspondences in a mosaicking context. By designing overlap models and an informativeness measure in a principled way, our framework is able to retrieve long-range pairs for arbitrarily long sequences. This theoretical property was verified in practice on synthetic trajectories. Moreover we also demonstrated the use of our framework on two practical application cases: the reduction of matches in bundle adjustment reconstruction, and the interactive monitoring of a mosaic reconstruction during fetal surgery. 

\begin{acknowledgements}
This work was partially supported by EPSRC [EP/L016478/1], Wellcome / EPSRC [203145Z/16/Z; NS/A000050/1; WT101957; NS/A000027/1]  and the ERC Starting Grant 677697. Jan Deprest is supported by the Great Ormond Street Hospital Charity. This work was undertaken at UCL and UCLH, which receive a proportion of funding from the DoH NIHR UCLH BRC funding scheme. We are grateful to George Attilakos and Ruwan Wimalasundera from UCLH for their help in the acquisition of the fetoscopy dataset.
\\
\textbf{Conflict of Interest } Tom Vercauteren owns shares from Mauna Kea Technologies.

\end{acknowledgements}


{\small
\bibliography{Mosaicking}   
}

\appendix

\section{Equivalence between Eq.~\ref{eq:affine_bundle_adjustment} and Eq.~\ref{eq:bundle_adjustment_affine_vectorised_first_line}}
\label{appendix:bundle_adjustment_affine_vectorised}

In this appendix, we derive detailed derivations showing that
\begin{equation}
(\hat{\boldsymbol{\Theta}}_n)_{n \neq r} = \argmin_{(\boldsymbol{\Theta}_n)_{n \neq r}} \sum_{(i,j) \in \mathcal{P}_k^+} \sum_{l=1}^{L_{ij}} \Vert \boldsymbol{\Theta}_j \tilde{\textbf{x}}^{(i,j)}_{j,l} - \boldsymbol{\Theta}_i  \tilde{\hat{\textbf{x}}}^{(i,j)}_{i,l} \Vert^2,
\tag{\ref{eq:affine_bundle_adjustment}}
\end{equation}
can be rewritten
\begin{equation}
\vecaff[(\hat{\boldsymbol{\Theta}}_n)_{n \neq r}] = \argmin_{\bm{\theta} \in \mathbb{R}^{6(N-1)}} \sum_{(i,j) \in \mathcal{P}_k^+} \Vert \bold{A}_{ij}^{\mathrm{T}} \bm{\theta} - \bold{b}_{ij} \Vert^2, \tag{\ref{eq:bundle_adjustment_affine_vectorised_first_line}}
\end{equation}
where
\begin{equation}
\bold{A}_{ij} = \bar{\delta}_{jr} \bold{e}_{j - \delta_{j>r}}^{(N-1)} \otimes  \bold{I}_2 \otimes \tilde{\bold{X}}^{(i,j)}_j 
 - \bar{\delta}_{ir} \bold{e}_{i - \delta_{i>r}}^{(N-1)} \otimes  \bold{I}_2 \otimes \tilde{\hat{\bold{X}}}^{(i,j)}_i
\tag{\ref{eq:definition_of_A}}
\end{equation}
and
\begin{equation}
\bold{b}_{ij} =  \delta_{ir} \vecrow\left(\hat{\bold{X}}^{(i,j)}_i\right) - \delta_{jr} \vecrow\left(\bold{X}^{(i,j)}_j\right).
\tag{\ref{eq:definition_of_b}}
\end{equation}
We consider a given pair $(i,j) \in \mathcal{P}_k^+$.  For a fixed $l \in \lbrace1, \ldots, L_{ij}\rbrace$, we have 
\begin{equation}
\Vert \boldsymbol{\Theta}_j \tilde{\textbf{x}}^{(i,j)}_{j,l} - \boldsymbol{\Theta}_i  \tilde{\hat{\textbf{x}}}^{(i,j)}_{i,l} \Vert = \Vert \bold{P}_{c \leftarrow h} \left( \boldsymbol{\Theta}_j \tilde{\textbf{x}}^{(i,j)}_{j,l} - \boldsymbol{\Theta}_i  \tilde{\hat{\textbf{x}}}^{(i,j)}_{i,l}\right) \Vert
\end{equation}
since the third coordinate of the $3 \times 1$ vector $\boldsymbol{\Theta}_j \tilde{\textbf{x}}^{(i,j)}_{j,l} - \boldsymbol{\Theta}_i  \tilde{\hat{\textbf{x}}}^{(i,j)}_{i,l}$ is always $0$ as the difference of two points expressed in homogeneous coordinates. We can also notice that the vector $\bold{P}_{c \leftarrow h} \left( \boldsymbol{\Theta}_j \tilde{\textbf{x}}^{(i,j)}_{j,l} - \boldsymbol{\Theta}_i  \tilde{\hat{\textbf{x}}}^{(i,j)}_{i,l}\right)$ is the $l^{\textrm{th}}$ column of the matrix $\bold{P}_{c \leftarrow h} \left( \boldsymbol{\Theta}_j \tilde{\bold{X}}^{(i,j)}_j - \boldsymbol{\Theta}_i  \tilde{\bold{X}}^{(i,j)}_i\right)$. As a result, the sum over $l$ can be compactly written 
\begin{multline}
\sum_{l=1}^{L_{ij}} \Vert \boldsymbol{\Theta}_j \tilde{\textbf{x}}^{(i,j)}_{j,l} - \boldsymbol{\Theta}_i  \tilde{\hat{\textbf{x}}}^{(i,j)}_{i,l} \Vert^2 = \\ \Vert \vecrow \left( \bold{P}_{c \leftarrow h}  \boldsymbol{\Theta}_j\tilde{\bold{X}}^{(i,j)}_j \right) - \vecrow \left( \bold{P}_{c \leftarrow h}  \boldsymbol{\Theta}_i  \tilde{\bold{X}}^{(i,j)}_i\right)  \Vert^2.
\label{eq:bundle_adjustment_vectorised_sum}
\end{multline}
To show that Eq.~\ref{eq:bundle_adjustment_affine_vectorised_first_line} holds, it is now sufficient to show that
\begin{multline}
\vecrow (\bold{P}_{c \leftarrow h}  \boldsymbol{\Theta}_j\tilde{\bold{X}}^{(i,j)}_j ) = \\
\bar{\delta}_{jr} [\bold{e}_{j - \delta_{j>r}}^{(N-1)} \otimes  \bold{I}_2 \otimes \tilde{\bold{X}}^{(i,j)}_j]^\mathrm{T} \vecaff[(\boldsymbol{\Theta}_n)_{n \neq r}] - \delta_{jr} \vecrow (\bold{X}^{(i,j)}_j)
\label{eq:bundle_adjustment_affine_vectorised_j}
\end{multline}
for any $j$. Indeed, using this equality for both $i$ and $j$ and introducing them in Eq.~\ref{eq:bundle_adjustment_vectorised_sum} immediately leads to Eq.~\ref{eq:bundle_adjustment_affine_vectorised_first_line}. The remainder of this appendix aims at proving Eq.~\ref{eq:bundle_adjustment_affine_vectorised_j}. For compactness, we denote $\tilde{\bold{X}}_j = \tilde{\bold{X}}^{(i,j)}_j$ from now on in this appendix, ignoring the dependency on $(i,j)$ in the notation.  By application of Eq.~\ref{eq:vectorise_product}, we have
\begin{equation}
\vecrow \left(\bold{P}_{c \leftarrow h}  \boldsymbol{\Theta}_j \tilde{\bold{X}}_j \right) = \left( \bold{P}_{c \leftarrow h} \otimes \tilde{\bold{X}}_j^\mathrm{T}\right)  \vecrow \left( \boldsymbol{\Theta}_j \right).
\end{equation}
Using Eq.~\ref{eq:vec_to_vec_6} stating that
\begin{equation}
\vecrow(\boldsymbol{\Theta}_j) =  (\bold{P}_{c \leftarrow h}^{\mathrm{T}}\otimes \bold{I}_3) \vecaff(\boldsymbol{\Theta}_j) + \bold{e}_{9}^{(9)},
\tag{\ref{eq:vec_to_vec_6}}
\end{equation}
we can use the multiplicative properties of the Kronecker product to obtain
\begin{multline}
\vecrow \left(\bold{P}_{c \leftarrow h}  \boldsymbol{\Theta}_j \tilde{\bold{X}}_j \right) = \\
\left[\left(\bold{P}_{c \leftarrow h} \bold{P}_{c \leftarrow h}^{\mathrm{T}}\right)  \otimes \tilde{\bold{X}}_j^\mathrm{T}\right] \vecaff(\boldsymbol{\Theta}_j) +  \left( \bold{P}_{c \leftarrow h} \otimes \tilde{\bold{X}}_j^\mathrm{T}\right)  \bold{e}_{9}^{(9)}.
\end{multline}
It is easy to verify that $\bold{P}_{c \leftarrow h} \bold{P}_{c \leftarrow h}^{\mathrm{T}} = \bold{I}_2$ and that $\left( \bold{P}_{c \leftarrow h} \otimes \tilde{\bold{X}}_j^\mathrm{T}\right)  \bold{e}_{9}^{(9)}$ is a vector composed of zeros due to the fact that the last column of $\bold{P}_{c \leftarrow h}$ is, itself, composed of zeros. Including these two results in the expression above leads to
\begin{equation}
\vecrow \left(\bold{P}_{c \leftarrow h}  \boldsymbol{\Theta}_j \tilde{\bold{X}}_j \right) = 
\left( \bold{I}_2 \otimes \tilde{\bold{X}}_j^\mathrm{T}\right) \vecaff(\boldsymbol{\Theta}_j).
\label{eq:derivation_ba_six_parameters}
\end{equation}
To relate the parameters $\vecaff(\boldsymbol{\Theta}_j)$ of the transformation $\boldsymbol{\Theta}_j$ to the entire vector of parameters $\vecaff[(\boldsymbol{\Theta}_n)_n]$ encoding the position of all the frames of the sequence, we use Eq.~\ref{eq:vec_6_to_vec_6_collection} stating that
\begin{equation}
\vecaff(\boldsymbol{\Theta}_j) = \left[\left( \bold{e}_j^{(N)}\right)^{\mathrm{T}} \otimes \bold{I}_6\right] \vecaff[(\boldsymbol{\Theta}_n)_n].
\tag{\ref{eq:vec_6_to_vec_6_collection}}
\end{equation}
To incorporate Eq.~\ref{eq:vec_6_to_vec_6_collection} into Eq.~\ref{eq:derivation_ba_six_parameters}, we write $\bold{I}_6 = \bold{I}_2 \otimes \bold{I}_3$ and, by associativity of the Kronecker product, $\left( \bold{e}_j^{(N)}\right)^{\mathrm{T}} \otimes \bold{I}_6$ can then be rewritten as $\left[ \left( \bold{e}_j^{(N)}\right)^{\mathrm{T}} \otimes \bold{I}_2 \right] \otimes \bold{I}_3$. By doing so, the products $\left[ \left( \bold{e}_j^{(N)}\right)^{\mathrm{T}} \otimes \bold{I}_2 \right] \otimes \bold{I}_3$ and $\left( \bold{I}_2 \otimes \tilde{\bold{X}}_j^\mathrm{T}\right)$ are of compatible dimensions such that the multiplicative property of the Kronecker product can be applied, leading to
\begin{equation}
\vecrow \left(\bold{P}_{c \leftarrow h}  \boldsymbol{\Theta}_j \tilde{\bold{X}}_j \right) = \left[ \left( \bold{e}_j^{(N)}\right)^{\mathrm{T}} \otimes \bold{I}_2 \otimes \tilde{\bold{X}}_j^\mathrm{T}\right] \vecaff[(\boldsymbol{\Theta}_n)_n].
\label{eq:derivation_ba_all_parameters}
\end{equation}
A final step must be conducted to reach the desired result. The vector $\vecaff[(\boldsymbol{\Theta}_n)_n]$ includes the reference frame, whereas the bundle adjustment formulation is parametrised by the vector of $6(N-1)$ parameters $\vecaff[(\boldsymbol{\Theta}_n)_{n \neq r}]$ instead, with $\boldsymbol{\Theta}_r$ being set as the identity. To express Eq.~\ref{eq:derivation_ba_all_parameters} in terms of $\vecaff[(\boldsymbol{\Theta}_n)_{n \neq r}]$, we define the $N \times (N-1)$ matrix
\begin{equation}
\bold{P}_r = 
\begin{pmatrix}
\bold{I}_{r - 1} & \textbf{O}_{(r-1) \times (N-r)} \\
\textbf{0}^{\mathrm{T}}_{r-1} & \textbf{0}^{\mathrm{T}}_{N-r}\\
\textbf{O}_{(N-r)\times (r-1)} & \bold{I}_{N-r}
\end{pmatrix},
\end{equation}
which allows us to write
\begin{equation}
\vecaff[(\boldsymbol{\Theta}_n)_n] = (\bold{P}_r \otimes \bold{I}_6) \vecaff[(\boldsymbol{\Theta}_n)_{n \neq r}] + \bold{e}_{r}^{(N)} \otimes \vecrow(\bold{P}_{c \leftarrow h}).
\label{eq:insert_reference_in_frame}
\end{equation}
Intuitively, the right hand-side of Eq.~\ref{eq:insert_reference_in_frame} extends the vector of unknown parameters $\vecaff[(\boldsymbol{\Theta}_n)_{n \neq r}]$ by inserting, at the right position, the known parameters $\vecaff(\bold{I}_3) = \vecrow(\bold{P}_{c \leftarrow h})$ of the transformation $\boldsymbol{\Theta}_r$. Using again the decomposition $\bold{P}_r \otimes \bold{I}_6 = \bold{P}_r \otimes \bold{I}_2 \otimes \bold{I}_3$, inserting Eq.~\ref{eq:insert_reference_in_frame} into Eq.~\ref{eq:derivation_ba_all_parameters} gives
\begin{multline}
\vecrow(\bold{P}_{c \leftarrow h} \boldsymbol{\Theta}_j \tilde{\bold{X}}_j) = [(\bold{P}_r^\mathrm{T} \bold{e}_j^{(N)}) \otimes \bold{I}_2 \otimes \tilde{\bold{X}}_j]^\mathrm{T} \vecaff[(\boldsymbol{\Theta}_n)_{n \neq r}] \\ 
+ \left[ \left( \bold{e}_j^{(N)}\right)^{\mathrm{T}} \bold{e}_{r}^{(N)} \right] \otimes \left[ \left( \bold{I}_2 \otimes \tilde{\bold{X}}_j^\mathrm{T}\right) \vecrow(\bold{P}_{c \leftarrow h}) \right].
\label{eq:derivation_ba_all_parameters_minus_one}
\end{multline}
The first term simplifies by noticing that $\bold{P}_r^{\mathrm{T}} \bold{e}_{j}^{(N)} = \bar{\delta}_{jr} \bold{e}_{j'}^{(N-1)}$ where $j' = j - \delta_{j > r}$. The second term simplifies by noticing that $(\bold{e}_j^{(N)})^{\mathrm{T}} \bold{e}_{r}^{(N)} = \delta_{jr}$ and that, using Eq.~\ref{eq:vectorise_product}, we have
\begin{equation}
\left( \bold{I}_2 \otimes \tilde{\bold{X}}_j^\mathrm{T}\right) \vecrow(\bold{P}_{c \leftarrow h})  = \vecrow(  \bold{I}_2 \bold{P}_{c \leftarrow h} \tilde{\bold{X}}_j) = \vecrow(\bold{X}_j).
\end{equation}
Including these final simplifications in Eq.~\ref{eq:derivation_ba_all_parameters_minus_one}, we have finally shown Eq.~\ref{eq:bundle_adjustment_affine_vectorised_j} and thus Eq.~\ref{eq:bundle_adjustment_affine_vectorised_first_line}.

\section{Computation of the Jacobian of $\boldsymbol{\gamma}_{ij}$}
\label{appendix:jacobian}
We detail here the computation of the Jacobian $\bold{J}_{ij}$ of 
\begin{equation}
\boldsymbol{\gamma}_{ij} = \bold{P}_{c \leftarrow h} \boldsymbol{\Theta}_{j}^{-1} \boldsymbol{\Theta}_i \tilde{\boldsymbol{\gamma}},
\tag{\ref{eq:definition_gamma_ij}}
\end{equation} 
seen as a function of the $12$ variables $\vecaff(\boldsymbol{\Theta}_i,\boldsymbol{\Theta}_j)$. This Jacobian is by definition the concatenation of the two $2 \times 6$ partial Jacobians $\bold{J}^{(i)}_{ij}$ and $\bold{J}^{(j)}_{ij}$ obtained by differentiation with respect to $\vecaff(\boldsymbol{\Theta}_i)$ and $\vecaff(\boldsymbol{\Theta}_j)$ respectively. 

We compute the first $2 \times 6$ block $\bold{J}^{(i)}_{ij}$ by differentiating with respect to $\vecaff(\boldsymbol{\Theta}_i)$. To do so, we notice that
\begin{align}
\boldsymbol{\gamma}_{ij} &= \vecrow(\boldsymbol{\gamma}_{ij}) \\
&= \vecrow(\bold{P}_{c \leftarrow h} \boldsymbol{\Theta}_{j}^{-1} \boldsymbol{\Theta}_i \tilde{\boldsymbol{\gamma}}) \\
&= [(\bold{P}_{c \leftarrow h}  \boldsymbol{\Theta}_j^{-1}) \otimes \tilde{\boldsymbol{\gamma}})^{\mathrm{T}}] \vecrow (\boldsymbol{\Theta}_i) \\
&= [(\bold{P}_{c \leftarrow h}  \boldsymbol{\Theta}_j^{-1}) \otimes \tilde{\boldsymbol{\gamma}}^{\mathrm{T}}] [(\bold{P}_{c \leftarrow h}^{\mathrm{T}} \otimes \bold{I}_3) \vecaff(\boldsymbol{\Theta}_i) + \bold{e}_9^{(9)}]\\
&= [(\bold{P}_{c \leftarrow h}  \boldsymbol{\Theta}_j^{-1} \bold{P}_{c \leftarrow h}^{\mathrm{T}}) \otimes \tilde{\boldsymbol{\gamma}}^{\mathrm{T}}] \vecaff(\boldsymbol{\Theta}_i) + C \\
&= [(\boldsymbol{\Theta}_j^{\textrm{(lin)}})^{-1}\otimes \tilde{\boldsymbol{\gamma}}^{\mathrm{T}}] \vecaff(\boldsymbol{\Theta}_i) + C,
\end{align}
where $C$ is constant with respect to $\boldsymbol{\Theta}_i$. The derivations above make use of the notations and properties introduced in Section~\ref{sec:notations}. As a result of this linear relationship in $\vecaff(\boldsymbol{\Theta}_i)$, it is immediate that the first $2 \times 6$ block of the Jacobian evaluated at $\vecaff(\hat{\boldsymbol{\Theta}}_i,\hat{\boldsymbol{\Theta}}_j)$ is 
\begin{equation}
\bold{J}^{(i)}_{ij} = (\hat{\boldsymbol{\Theta}}_j^{\textrm{(lin)}})^{-1}\otimes \tilde{\boldsymbol{\gamma}}^{\mathrm{T}}.
\end{equation}
We have just proved Eq.~\ref{eq:jacobian_gamma_i}.

To compute the second block $\bold{J}^{(j)}_{ij}$, we consider the derivative of $\boldsymbol{\gamma}_{ij}$ with respect to a parameter $\theta_{d}$ of $\boldsymbol{\Theta}_j$, with $d \in \left\lbrace 1, \ldots, 6\right\rbrace$. This quantity is by definition the $d^{\textrm{th}}$ column of $\bold{J}^{(j)}_{ij}$. and is given as
\begin{align}
\frac{\partial \boldsymbol{\gamma}_{ij}}{\partial \theta_{d}} &= - \bold{P}_{c \leftarrow h} \boldsymbol{\Theta}_{j}^{-1} \frac{\partial \boldsymbol{\Theta}_j}{\partial \theta_{d}} \boldsymbol{\Theta}_{j}^{-1} \boldsymbol{\Theta}_i \tilde{\boldsymbol{\gamma}} \label{eq:derivative_inverse_j}\\ 
&= - \bold{P}_{c \leftarrow h} \boldsymbol{\Theta}_{j}^{-1} \frac{\partial \boldsymbol{\Theta}_j}{\partial \theta_{d}} \tilde{\boldsymbol{\gamma}}_{ij}.
\end{align}
In Eq.~\ref{eq:derivative_inverse_j}, we used the general identity 
\begin{equation}
\frac{\partial \bold{M}(x)^{-1}}{\partial x} =  - \bold{M}(x)^{-1} \frac{\partial \bold{M}(x)}{\partial x} \bold{M}(x)^{-1}
\end{equation}
giving the derivative of each coefficient of the inverse of a matrix $\bold{M}$.
Similarly as above, we can then write
\begin{align}
\frac{\partial \boldsymbol{\gamma}_{ij}}{\partial \theta_{d}} &= \vecrow\left( \frac{\partial \boldsymbol{\gamma}_{ij}}{\partial \theta_{d}}\right)\\
&= - \left[ \left( \bold{P}_{c \leftarrow h} \boldsymbol{\Theta}_{j}^{-1}\right) \otimes \tilde{\boldsymbol{\gamma}}_{ij}^\mathrm{T}\right] \vecrow\left( \frac{\partial \boldsymbol{\Theta}_j}{\partial \theta_{d}} \right) \\
&= - \left[ \left( \bold{P}_{c \leftarrow h} \boldsymbol{\Theta}_{j}^{-1}\right) \otimes \tilde{\boldsymbol{\gamma}}_{ij}^\mathrm{T}\right] \left[\bold{P}_{c \leftarrow h}^{\mathrm{T}} \otimes \bold{I}_3 \right] \vecaff\left( \frac{\partial \boldsymbol{\Theta}_j}{\partial \theta_{d}} \right) \\
&= - \left[ \left( \bold{P}_{c \leftarrow h} \boldsymbol{\Theta}_{j}^{-1} \bold{P}_{c \leftarrow h}^{\mathrm{T}}\right) \otimes \tilde{\boldsymbol{\gamma}}_{ij}^\mathrm{T}\right] \bold{e}^{(6)}_d \\
&= - \left[(\boldsymbol{\Theta}_j^{\textrm{(lin)}})^{-1} \otimes \tilde{\boldsymbol{\gamma}}_{ij}^\mathrm{T} \right] \bold{e}^{(6)}_d.
\end{align}
The final quantity, via the multiplication by $\bold{e}^{(6)}_d$, extracts the $d^{\textrm{th}}$ column of $(- (\boldsymbol{\Theta}_j^{\textrm{(lin)}})^{-1} \otimes \tilde{\boldsymbol{\gamma}}_{ij}^\mathrm{T})$. Since this quantity is also the $d^{\textrm{th}}$ column of the second block of the Jacobian, we can conclude that the latter is equal to 
\begin{equation}
\bold{J}^{(j)}_{ij} = -(\hat{\boldsymbol{\Theta}}_j^{\textrm{(lin)}})^{-1} \otimes \tilde{\hat{\boldsymbol{\gamma}}}_{ij}^\mathrm{T}.
\end{equation}
This proves Eq.~\ref{eq:jacobian_gamma_j}.

\end{document}